\documentclass[final,12pt]{clear2023} % Include author names

\title[Jointly Learning Causal Abstractions With Multiple Interventions]{Jointly Learning Consistent Causal Abstractions Over Multiple Interventional Distributions}
\usepackage{times}
\usepackage{tikz}
\usepackage{adjustbox}
\usepackage{xcolor}
\usepackage{float}
\usetikzlibrary{positioning}
\usetikzlibrary{arrows.meta}
\definecolor{mygray}{gray}{0.85}
\usepackage{tikz-cd}
\usepackage{multicol,caption}
\usepackage{stmaryrd}
\usepackage{float}
\usepackage{array}
\usepackage{algorithmicx}
\usepackage{algpseudocode}
\usepackage{algorithm}
\usepackage{mathrsfs}
\usepackage{mathtools}
\usepackage{capt-of}
\usepackage{enumitem}
\usepackage{multicol}

\newcommand{\scm}[1]{\mathcal{M}#1}
\newcommand{\dgraph}{\mathcal{G}}
\newcommand{\abs}{\boldsymbol{\alpha}}

\newcommand{\iiks}[2]{\left\{ \begin{array}{c}
#1\\
#2
\end{array}\right\}}
\newcommand{\tmp}[1]{t_{col}(#1)}

\newcommand{\inlineiiks}[2]{\left\{ \begin{smallmatrix}#1 \\ #2 \end{smallmatrix}\right\}}
\newcommand{\lone}{\ell_1}

\setitemize{itemsep=-5pt}
\setenumerate{itemsep=-2pt}

\tikzset{
mynode/.style={
  draw,
  circle,
  text width=0.6cm,
  minimum size=0.6cm,
  align=center,
  fill=mygray
  }
}

\tikzset{
encoden/.style={
  draw,
  circle,
  text width=0.6cm,
  minimum size=0.6cm,
  align=center
  }
}

\tikzset{
mytext/.style={
  text width=0.6cm,
  minimum size=0.6cm,
  align=center
  }
}

\tikzset{
mysimple/.style={
  draw,
  circle,
  fill=mygray
  }
}

\tikzset{
arr/.style = {-{Triangle[length=2.5mm, width=2mm]}}
}
\tikzset{
arr1/.style = {->,line width=1pt}
}

\clearauthor{\Name{Fabio Massimo Zennaro} \Email{fabio.zennaro@warwick.ac.uk}\\
\Name{Máté Drávucz} \Email{mate.dravucz@warwick.ac.uk}\\
\Name{Geanina Apachitei} \Email{geanina.apachitei@warwick.ac.uk}\\
\Name{W. Dhammika Widanage} \Email{dhammika.widanalage@warwick.ac.uk}\\
\Name{Theodoros Damoulas} \Email{t.damoulas@warwick.ac.uk}\\
\addr Dept. of Computer Science \& Dept. of Statistics \& WMG, University of Warwick, Coventry, CV4 7AL, UK\\
\addr The Faraday Institution, Harwell Science and Innovation, Campus, Quad One, Didcot, UK}

\begin{document}
\maketitle

\begin{abstract}%
%Structural causal models (SCMs) are an accepted formalism to represent causal systems in isolation. Causal models, however, can be related among themselves by a relationship of abstraction. 
An abstraction can be used to relate two structural causal models representing the same system at different levels of resolution.
%Given two SCMs, 
Learning abstractions which guarantee consistency with respect to interventional distributions would allow one to jointly reason about evidence across multiple levels of granularity while respecting the underlying cause-effect relationships. In this paper, we introduce a first framework for causal abstraction learning between SCMs based on the formalization of abstraction recently proposed by \cite{rischel2020category}. 
%and characterize the asymptotic complexity of the associated combinatorial optimization problem. 
Based on that, we propose a differentiable programming solution that jointly solves a number of combinatorial sub-problems, and we study its performance and benefits against independent and sequential approaches on synthetic settings and on a challenging real-world problem related to electric vehicle battery manufacturing.
\end{abstract}


\section{Introduction}

\begin{figure}[H]
\begin{adjustbox}{center}
\begin{tikzpicture}[shorten >=1pt, auto, node distance=1cm, thick, scale=0.8, every node/.style={scale=0.8}]
	\tikzstyle{node_style} = [circle,draw=black]
	\node[node_style] (S) at (0,0) {S};
	\node[node_style] (T) at (2,0) {T};
	\node[node_style] (C) at (4,0) {C};
	\draw[->]  (S) to (T);
	\draw[->]  (T) to (C);
	\end{tikzpicture}
\hspace{3cm}
\begin{tikzpicture}[shorten >=1pt, auto, node distance=1cm, thick, scale=0.8, every node/.style={scale=0.8}]
	\tikzstyle{node_style} = [circle,draw=black]
	\node[node_style] (S) at (0,0) {S'};
	\node[node_style] (C) at (2,0) {C'};
	\draw[->]  (S) to (C);
\end{tikzpicture}
\end{adjustbox}
\caption{Lung cancer SCM designed by lab $L$ (left) and lab $L'$ (right).} \label{fig:SCMs}
\end{figure}
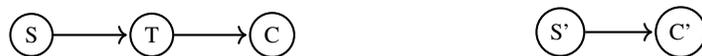

Causal models are conceptual constructs we use in our everyday understanding of the world and in scientific modelling. Structural causal models (SCM) provide a mathematical formalism to express causal assumptions, encode quantities of interest, and reason about relationships of cause and effect. For instance, a research lab $L$ may investigate lung cancer, and decide to model this scenario considering causal connections between a set of relevant variables, such as the smoking habits of patients, the presence of tar deposits in their lungs, and whether they ended up developing lung cancer (see Figure \ref{fig:SCMs}(left)). 
Another common feature of reasoning and scientific modelling is reliance on multiple levels of abstraction, whereby an identical system can be studied at multiple levels of granularity. For instance, in studying lung cancer, another research lab $L'$ may decide to record only two variables, ignoring the contribution of tar deposits (see Figure \ref{fig:SCMs}(right)).

While SCMs allow us to deal with causal relationships internal to a given model, an abstraction focuses on \textit{external} relations between different models. The idea of abstraction is widespread in artificial intelligence: it underlies the very notion of intelligence \citep{mitchell2021abstraction}, it has been invoked to explain the success of deep learning \citep{lecun2015deep}, and it has a central role in causal representation learning \citep{scholkopfLBKKGB21}. However, rigorous formalisms for abstractions have only been recently proposed \citep{rubenstein2017causal,beckers2019abstracting,rischel2020category}.

Our work starts from the abstraction framework of \cite{rischel2020category}, which provides a grounded way to express an abstraction between two SCMs and to quantitatively assess its consistency. Evaluation of consistency requires, beyond the definition of the SCMs, a full specification of an abstraction, which, in reality, may rarely be available. In this paper, we consider the problem of \textit{learning an abstraction} when only partial information about it is available. This would correspond, for instance, to the case in which two labs are aware of their own SCMs but do not have an exact mapping between them. Successfully learning an abstraction would enable them to automatically transfer data and results across the models; in a low-data regime, where collecting new samples may be costly, having a proper abstraction would allow them to integrate their evidence and improve inferences.

%In this paper we introduce the problem of learning the complete specification of an abstraction when two SCMs are given and little information about the abstraction is provided. Solving such a problem, would allow for models at different levels of abstraction to be properly aligned and for an automatic transfer of results and data across them. For instance, provided with a correct abstraction, researchers from the lab $M'$ may abstract data produced by the researchers from the lab $M$, integrate these abstracted data with their own data, and then perform inferences or prediction relying on a larger amount of data.

Our contributions are introducing a new learning problem with specific semantics (abstraction learning) but a very generic syntax (commutativity learning); proposing a relaxation of the ensuing combinatorial problem and a solution based on differentiable programming which jointly solves a number of combinatorial sub-problems at once; analyzing empirically the performance of this approach on synthetic settings; demonstrating the benefits of our approach on the important problem of learning coating models for the batteries of electric vehicles (EV) by learning an abstraction that allows us to relate small-scale datasets collected through expensive real-world experiments performed in labs across France and the UK. To the best of our knowledge, this work also constitutes the first real-world application of an abstraction learning framework between SCMs.

The rest of the paper is organized as follows. Section \ref{sec:Background} reviews important background definitions, and Section \ref{sec:RelatedWork} presents related work. Section \ref{sec:ProblemStatement} discusses the learning problem. Section \ref{sec:Methodology} introduces our proposed methodology, 
%from the relaxation of the problem and to the final algorithm based on differentiable programming. 
and Section \ref{sec:Experiments} presents our empirical results. Section \ref{sec:Discussion} summarizes our work and offers considerations on our approach and results. Appendix \ref{app:Notation} offers a summary of the notation used throughout this paper.

\section{Background} \label{sec:Background}
In this section we provide definitions for the main concepts related to causality  and abstraction.

\subsection{Causality}\label{ssec:Causality}

In order to express causal models, we rely on Pearl's formalism of structural causal models \citep{pearl2009causality,peters2017elements}. See \cite{pearl2009causality} for the analogies and differences to the potential outcomes framework by Rubin \citep{rubin2005causal}. %which allows us to express causal assumptions in a graphical form and evaluates quantitatively causal relationships in a probabilistic form.

\begin{definition}[Structural Causal Model (SCM) \citep{pearl2009causality}]
A structural causal model $\scm{}$ consists of a directed acyclic graph (DAG) $\dgraph_{\scm{}}$, and a tuple $\langle \mathcal{U}, \mathcal{X}, \mathcal{F}, P(\mathcal{U})\rangle$, where:
\begin{itemize}
    \item $\mathcal{U}$ is a finite set of exogenous (unobserved or latent) variables;
    \item $\mathcal{X}$ is a finite set of endogenous (observed) variables, each one with a discrete domain $\mathcal{M}[X_i]$; 
    \item $\mathcal{F} = \{f_{1}, . . . , f_{|\mathcal{X}|}\}$ is a set of modular functions (mechanisms) such that $x_i = f_i (pa(X_i))$, where $x_i$ is the value of an endogenous variable $X_i \in \mathcal{X}$ and $pa(X_i) \subseteq (\mathcal{U} \cup \mathcal{X}) \setminus X_i$.
    \item $P(\mathcal{U})$ is a joint distribution over the exogenous variables.
\end{itemize}

\end{definition}

Notice that the assumptions of acyclicity and joint distribution over the exogenous variables imply a \emph{semi-Markovian SCM}; in this setting, the set $pa(X_i)$ of variables that determines the value of an endogenous variable $X_i$ can be given the graph-theoretic interpretation of \emph{parents} of $X_i$, and edges can be interpreted as causal relations. The additional assumption of discrete domains for each endogenous variable is introduced as a requirement of the abstraction framework of \cite{rischel2020category}.
Models presented in Figure \ref{fig:SCMs} can be given a rigorous reading as DAGs $\mathcal{G}_{\scm{}}$ and $\mathcal{G}_{\scm{'}}$.
%underlying two different SCMs $\scm{}$ and $\scm{'}$.

A SCM allows us to study a system not only in the observational domain, but also under interventions. Modularity of the mechanisms allows us to formally define interventions.

\begin{definition}[Intervention \citep{pearl2009causality}]
Given a SCM $\scm{}=\langle \mathcal{U}, \mathcal{X}, \mathcal{F}, P(\mathcal{U})\rangle$, an intervention $\iota \coloneqq do(X_i = x_i)$ is an operator that generates a new post-interventional SCM $\scm{_\iota}=\langle \mathcal{U}_\iota, \mathcal{X}_\iota, \mathcal{F}_\iota, P_\iota(\mathcal{U}_\iota)\rangle$ where $\mathcal{U}_\iota = \mathcal{U}$,
$\mathcal{X}_\iota = \mathcal{X}$,
 $P_\iota(\mathcal{U}_\iota)=P(\mathcal{U})$, while
 $f_{\iota_i} = x_i$ and $f_{\iota_j} = f_j$ for all $j \neq i$.
\end{definition}

Thus, an intervention $\iota \coloneqq do(X_i = x_i)$ creates a new model $\scm{_\iota}$, identical to the original one except for the structural function $f_i$ which is replaced with the constant $x_i$; the DAG of the post-interventional model $\scm{_\iota}$ is similarly identical, with the exception of the node associated with $X_i$ which has all the incoming edges removed. It is immediate to extend the definition of intervention to multiple endogenous variables: $do(\mathbf{X} = \mathbf{x})$, where $\mathbf{X} = [X_1, X_2,..., X_m]$ is a vector of variables in $\mathcal{X}$ and $\mathbf{x} = [x_1, x_2,..., x_m]$ is a vector of values associated with each variable.

\subsection{Abstraction}\label{ssec:Abstraction}

We introduce a notion of abstraction meant to relate two SCMs representing an identical system. This definition originates from category theory, and it assumes SCMs with a finite set $\mathcal{X}$ of variables, each one defined on a finite and discrete domain $\mathcal{M}[X_i]$ \citep{rischel2020category}.

\begin{definition}[Abstraction \citep{rischel2020category}]
Given two SCMs $\scm{}$ and $\scm{'}$, an abstraction $\abs$ is a tuple $\langle {R}, {a}, \alpha \rangle$ where:
\begin{itemize}
    \item $R \subseteq \mathcal{X}$ is a subset of relevant variables in the model $\mathcal{M}$;
    \item ${a} : R \rightarrow \mathcal{X'}$ is a surjective map between variables, from nodes in $\mathcal{M}$ to node in $\mathcal{M'}$;
    \item $\alpha$ is a collection of surjective maps $\alpha_{X'}: \mathcal{M}[{a}^{-1}(X')] \rightarrow \mathcal{M'}[X']$ from outcomes of variables in $\mathcal{M}$ to outcomes of variables in $\mathcal{M'}$.
\end{itemize}
\end{definition}

An abstraction establishes an asymmetric relation from a base or low-level model $\scm{}$ to an abstracted or high-level model $\scm{'}$. This definition encodes a mapping on two layers: on a structural or graphical level between the nodes of the DAGs via $a$, and on a distributional level via the maps $\alpha_{X'}$ \citep{zennaro2022abstraction}. 
%However, this definition does not capture in itself the meaning of a well-behaved abstraction.
%Informally, we say that the two model are in a relation of abstraction if the behave in a consistent way. 
Since we are dealing with causal models, we require the SCMs to behave consistently wrt interventions. 
%\theo{These last two sentences could be ommited/compressed if in need of space}
%Let us formalize this intuition.

\begin{definition}[Consistency wrt an interventional distribution]\label{def:AbstractionConsistency}
Let $\abs$ be an abstraction from $\scm{}$ to $\scm{'}$. Let $\mathbf{X'}$ and $\mathbf{Y'}$ be two disjoint subsets of variables in $\mathcal{X'}$. The abstraction $\abs{}$ is consistent wrt the interventional distribution $P'(\mathbf{Y'} \vert do(\mathbf{X'}))$ if the following diagram commutes:

\begin{adjustbox}{center}
\begin{tikzpicture}[thin,,scale=0.8, every node/.style={scale=0.8}]
\node[] (v1){$\mathcal{M}[\mathrm{a}^{-1}(\mathbf{X'})]$};
\node[right= 1.5cm of v1] (v2){$\mathcal{M}[\mathrm{a}^{-1}(\mathbf{Y'})]$};
\node[below= 1.5cm of v1] (v3){$\mathcal{M'}[\mathbf{X'}]$};
\node[below= 1.5cm of v2] (v4){$\mathcal{M'}[\mathbf{Y'}]$};
\draw[arr1] (v1) -- (v2); 
\draw[arr1] (v1) -- (v3);
\draw[arr1] (v2) -- (v4);
\draw[arr1] (v3) -- (v4);
\path [] (v1) -- node [midway,above,align=center ] {$\mu$} (v2);
\path [] (v1) -- node [midway,left,align=center ] {$\alpha_{\mathbf{X'}}$} (v3);
\path [] (v2) -- node [midway,right,align=center ] {$\alpha_{\mathbf{Y'}}$} (v4);
\path [] (v3) -- node [midway,below,align=center ] {$\nu$} (v4);
\end{tikzpicture}
\end{adjustbox}
that is:
\begin{equation}\label{eqn:commutativity}
    \alpha_{\mathbf{Y'}} \circ \mu = \nu \circ \alpha_{\mathbf{X'}},
\end{equation}
where $\mu$ and $\nu$ are the interventional distributions $P(a^{-1}(\mathbf{Y'}) \vert do(a^{-1}(\mathbf{X'})))$ and $P'((\mathbf{Y'}) \vert do((\mathbf{X'})))$.
%, respectively.

\end{definition}

Intuitively, commutativity means that we would obtain equivalent interventional results in two different ways: (i) by intervening on the base model, obtaining the interventional distribution $\mu$ on the base model and then abstracting via $\alpha_{\mathbf{Y'}}$; or, (ii) by intervening on the base model, abstracting via $\alpha_{\mathbf{X'}}$ and then obtaining the interventional distribution $\nu$ on the abstracted model.
Formally, commutativity has a category-theoretic meaning in the category of $\mathtt{FinStoch}$ where objects are sets and arrows are stochastic matrices \citep{fritz2020synthetic}. A rigorous explanation is provided in \cite{rischel2020category}, but here it is worth remarking that, when working with SCMs with finite variables and finite domains, every discrete variable (e.g.: ${X'}$) is associated with its domain set (e.g.: $\mathcal{M'}[{X'}]$), discrete distributions (e.g.: $P'(Y' \vert do(X'))$) can be encoded in stochastic matrices (e.g.: $\mu$), and deterministic abstractions (e.g.: $\alpha_{X'}$) can also be represented as binary stochastic matrices. 
Given this interpretation, the commutativity equality in Equation \ref{eqn:commutativity} boils down to an equality between matrix products. More details on this algebraic encoding are offered in Appendix \ref{app:AlgebraicEncoding}.
%Thus, all the nodes in the above diagram can be read as sets, and all the arrows as stochastic matrices; the commutativity equality in Equation \ref{eqn:commutativity} boils down to an equality between matrix products; in this way, the entire problem is given an algebraic interpretation. More details on this algebraic encoding are offered in Appendix \ref{app:AlgebraicEncoding}.\theo{Also here last 2-3 sentences could be ommited/compressed, moved to appendix if needed.}

As abstractions normally introduce approximations and rarely guarantee perfect commutativity,
%If commutativity does not hold, it is useful to have a measure of the approximation introduced by an abstraction. Again, 
it is useful to define an abstraction error wrt to interventions. 
%to measure such approximation.\theo{Can compress these two sentences into 1 sentence likely.}
\begin{definition}[Abstraction error wrt an interventional distribution \citep{rischel2020category}]\label{def:AbstractionError}
Let $\abs$ be an abstraction from SCM $\scm{}$ to SCM $\scm{'}$. Let $\mathbf{X'}$ and $\mathbf{Y'}$ be two disjoint subsets of variables in $\mathcal{X'}$. The abstraction error $E_{\abs}(\mathbf{X'},\mathbf{Y'})$ wrt the interventional distribution $P'(\mathbf{Y'} \vert do(\mathbf{X'}))$ is the Jensen-Shannon distance (JSD) between the upper and the lower path in the diagram in Definition \ref{def:AbstractionConsistency}:
\begin{equation}\label{eqn:abstractionerror}
    D_{JSD}(\alpha_{\mathbf{Y'}} \circ \mu; \nu \circ \alpha_{\mathbf{X'}}).
\end{equation}
\end{definition}

A definition of JSD is recalled in Appendix \ref{app:JSD}. Intuitively, the abstraction error quantifies the discrepancy between the upper and lower path in the abstraction diagram: how different are the results when (i) we first work with the low-level model and then abstract, and (ii) we first abstract and then work with the high-level model. This measure and the choice of JSD have a category-theoretic justification in the category $\mathtt{FinStoch}$ enriched in the category $\mathtt{Met}$ \citep{rischel2020category}.

%the choice of this metric is justified by \cite{rischel2020category} on the ground that it allows bounding the abstraction error whenever composing multiple abstractions. 
%Formally, this evaluation too has a category-theoretic meaning in the category $\mathtt{FinStoch}$ enriched in the category $\mathtt{Met}$ \citep{rischel2020category}.\theo{Most of this paragraph is repetition from Rischel so can be removed/compressed to just point to the reader to ref.}

From the above definitions, we can derive an overall notion of error.

\begin{definition}[Abstraction error] \label{def:OverallAbstractionError}
    Let $\abs$ be an abstraction from $\scm{}$ to $\scm{'}$. Let $\mathcal{J}$ be the set of all disjoint pair sets $(\mathbf{X'},\mathbf{Y'}) \in \mathscr{P}(\mathcal{X'})\times\mathscr{P}(\mathcal{X'}), \mathbf{X'} \cap \mathbf{Y'} = \emptyset$, where $\mathscr{P}()$ is the powerset. The abstraction error is:
    $$
        e(\abs) = \sup_{(\mathbf{X'},\mathbf{Y'}) \in \mathcal{J}} E_{\abs}(\mathbf{X'},\mathbf{Y'}).
    $$
    
\end{definition}

Thus, the overall abstraction error is simply the worst-case abstraction error when considering all possible interventional distributions. While $\mathcal{J}$ is formally defined as the set of all disjoint pair sets, it is possible to reduce $\mathcal{J}$ only to those pair sets $(\mathbf{X'},\mathbf{Y'})$ representing meaningful or relevant interventions $P'(\mathbf{Y'}\vert do(\mathbf{X'}))$. A consistent abstraction is then simply a zero-error abstraction:

\begin{definition}[Consistent abstraction]
    Let $\abs$ be an abstraction from $\scm{}$ to $\scm{'}$. The abstraction is consistent if, for all pairs $(\mathbf{X'},\mathbf{Y'})$ in $\mathcal{J}$, the abstraction $\abs{}$ is consistent wrt $P'(\mathbf{Y'} \vert do(\mathbf{X'}))$.
\end{definition}

\section{Related Work} \label{sec:RelatedWork}

Alternative accounts of abstraction have been offered in the literature. A seminal definition proposed by \cite{rubenstein2017causal}, and refined by \cite{beckers2019abstracting}, characterized abstraction only on a distributional level; measures of abstraction approximation in this context are discussed in \cite{beckers2020approximate}.
More detailed definitions that consider both the structural and the distributional levels, and that are grounded in category theory, have been presented in \cite{rischel2020category,rischel2021compositional,otsuka2022equivalence}. A review and a comparison of these definitions is offered in \cite{zennaro2022abstraction}. 
%In this work, we build over the work of \cite{rischel2020category}, but we do not provide novel contributions to the framework.
An attempt at defining a hierarchy of abstraction learning problems has been put forward in \cite{zennaro2022towards}.
%That work identified a hierarchy of learning problems and discussed potential loss functions, while this paper rigorously defines a precise learning problem, analyzes it, suggests a relaxation, and proposes an algorithm to solve it\theo{I would not compare against this work here (``while this paper.." - probably just state that only defined hierarchies etc without offering any methodology etc.}.\theo{compress this para and unite it with the last sentence from previous one since its work from Zennaro etc. that can easily be grouped}

Causal representation learning (CRL) \citep{chalupka2017causal} shares with this work a similar objective, but a different setup. Instead of learning a mapping between two SCMs, CRL normally starts from unstructured data and aims at learning causally-relevant representations. While abstraction learning as we defined it deals with mappings between SCMs, CRL may be seen as a preliminary or complementary step to abstraction learning, dealing with a mapping from an unstructured data space onto the space of causal variables potentially belonging to a SCM.

\section{Problem Statement}\label{sec:ProblemStatement}
The abstraction framework in Section \ref{sec:Background} provides a rigorous way to estimate an abstraction error once we are given a fully defined abstraction $\abs = \langle R,a,\alpha \rangle$ from $\scm{}$ to $\scm{'}$. 
Instead, we consider the scenario in which two SCMs $\scm{}$ and $\scm{'}$ are known, but only a partially defined abstraction in the form $\abs = \langle R,a \rangle$ is available. This represents the common situation where structural knowledge ($R,a$) is readily available, but detailed distributional knowledge ($\alpha$) lacking;
%we know what low-level variables are relevant ($R$), and how variables should be related ($a$), but we have no knowledge of how the outcomes of these variables should be mapped to each other ($\alpha$). 
in our lab example, this corresponds to the case where researchers from lab $L$ and $L'$ can exchange their models, agree on which variables are relevant, but they have no immediate knowledge on how to relate the results of their interventional experiments.
%\theo{rephrase to compress above para}
%This problem correspond to the \emph{completion/fixing problem} identified in the hierarchy of abstraction learning problems in \cite{zennaro2022towards}. 

Given two SMCs $\scm{}$ to $\scm{'}$, and a partial abstraction $\abs = \langle R,a \rangle$, abstraction learning is the problem of learning the values for the maps $\alpha_{X'}$ that minimize the abstraction error, achieving, if possible, a consistent abstraction. We can then cast the abstraction learning problem as an optimization problem:
\begin{equation}\label{eqn:optimizationProblem}
    \min_{\alpha\in\mathcal{A}}e(\boldsymbol{\alpha}),
\end{equation}
where $\mathcal{A}$ is the space of all feasible solutions for the collection of surjective maps $\alpha_{X'}$.
This optimization problem is challenging for three reasons: (i) it implies multiple sub-problems; (ii) these sub-problems are related; (iii) the solution space is combinatorial. %Let us analyze analyze each one of these aspects. 

\paragraph{Multiple sub-problems.}
Let us consider and unpack the optimization in Equation \ref{eqn:optimizationProblem}:
\begin{eqnarray}\label{eqn:SubProblems}
&\min_{\alpha\in\mathcal{A}} & e(\boldsymbol{\alpha}) = \\ &\min_{\alpha\in\mathcal{A}} &  \sup_{(\mathbf{X'},\mathbf{Y'}) \in \mathcal{J}} E_{\abs}(\mathbf{X'},\mathbf{Y'}) = \\
&\min_{\alpha\in\mathcal{A}} &  \sup \left\{  E_{\abs}(\mathbf{X'},\mathbf{Y'}), E_{\abs}(\mathbf{Y'},\mathbf{Z'}),
E_{\abs}(\mathbf{X'},\mathbf{Z'}) ...\right\} = \\
&\min_{\alpha\in\mathcal{A}} &  \sup \left\{ 
D_{JSD}(\alpha_{\mathbf{Y'}}  \mu; \nu  \alpha_{\mathbf{X'}}),
D_{JSD}(\alpha_{\mathbf{Z'}}  \mu'; \nu'  \alpha_{\mathbf{Y'}}),
D_{JSD}(\alpha_{\mathbf{Z'}}  \mu''; \nu''  \alpha_{\mathbf{X'}}) \label{eqn:SubProblems_lastline}
...\right\}.
\end{eqnarray}
The minimization of a supremum implies a minimization over multiple \emph{sub-problems}. Each sub-problem is defined by a pair set of endogenous variables $(\mathbf{X'},\mathbf{Y'}) \in \mathcal{J}$ representing the interventional distribution $P'(\mathbf{Y'} \vert do(\mathbf{X'}))$. For each interventional distribution we set up a diagram as in Definition \ref{def:AbstractionConsistency}, and we solve it in $\alpha_{\mathbf{X'}}$ and $\alpha_{\mathbf{Y'}}$ with the objective of minimizing the error $E_{\abs}(\mathbf{X'},\mathbf{Y'})$. Thus, we have a number of sub-problems equal to the cardinality $|\mathcal{J}|$, each one requiring the minimization of a JSD, as shown in Equation \ref{eqn:SubProblems_lastline}. Notice the direct correspondence between \emph{one} sub-problem, \emph{one} diagram, and \emph{one} minimization of a JSD.

\paragraph{Related sub-problems.}
The sub-problems identified above are not necessarily independent. As soon as we consider two interventional distributions involving one identical subset of variables, we will obtain two diagrams sharing an abstraction variable, For instance, if we consider $P'({Y'} \vert do({X'}))$ and $P'({Z'} \vert do({X'}))$, then the induced diagrams will share the abstraction map $\alpha_{{X'}}$; %Thus, the sub-problems can not be trivially solved in an independent fashion; in our example, 
this implies that minimizing the abstraction error for $E_{\abs}({X'},{Y'})$ by changing the value of $\alpha_{{X'}}$ will affect the abstraction error of $E_{\abs}({X'},{Z'})$ too.

\paragraph{Combinatorial optimization.}
The domain of each $\alpha_{X'}$, encoded as a binary stochastic matrix with shape $N_i \times M_i$, is $\{0,1\}^{N_{i}\times M_{i}}$, together with the constraint of stochasticity and surjectivity. There exists a finite number of solutions, equal to all possible surjective functions from a discrete $M_i$-dimensional space to a discrete $N_i$-dimensional space, which can be computed as
$
    N! \inlineiiks{M}{N},
$
where $\inlineiiks{M}{N} = \frac{1}{N!} \sum_{i=0}^{N} (-1)^{N-i} \binom{N}{i} i^M$ is the second-kind Stirling number \citep{graham1994concrete}. Consequently, since $\alpha$ is a collection of $|\mathcal{X}'|$ surjective functions, the space $\mathcal{A}$ of feasible solutions is the Cartesian product
%\begin{equation}\label{eqn:DiscreteSolutionSpace}
$    \mathcal{A} = \prod_{i=1}^{|\mathcal{X}'|}\{0,1\}^{N_{i}\times M_{i}}$,
%\end{equation}
with the constraint of stochasticity and surjectivity. The number of solutions, given by all possible combination of $\alpha_{X'}$ matrices, is:
\begin{equation}\label{eqn:all_n_surj_functions}
    |\mathcal{A}| = \prod_{i=1}^{|\mathcal{X}'|} N_i! \iiks{M_i}{N_i},
\end{equation}

The finiteness of the space $\mathcal{A}$ allows, theoretically, for searching an optimal solution by \emph{enumeration}. However, this quickly becomes unfeasible as the dimensionality of the input models grows.%\theo{verb missing above - rephrase and compress for hanging word at end}
%the worst-case computational complexity of finding a solution by enumeration, as a function of the dimensionality of the input SCMs, scales as %the dimensionality of $\mathcal{A}$; in the worst case this is super-exponential: $\left( \left( \frac{M}{\log M}! \exp^{M \log M} \right)^k \right)$. 
%A derivation is provided in Appendix \ref{app:AsymptoticComplexity}. 
%Next we will introduce a heuristics to compute (locally) optimal abstractions.
%While we will rely on enumeration to compute the exact ground-truth solution in low-dimensional synthetic experiments, we will next introduce a heuristics to compute optimal abstractions.

\section{Methodology} \label{sec:Methodology}
% explain our reasoning of neural net.

As the enumeration strategy is not generally feasible, some form of heuristic
%\theo{approximation heuristic?} 
becomes necessary. 
In this section we first describe our solution approach to the abstraction learning problem
%, as well as possible extensions. 
%\subsection{Solution approach}
%We explain our solution 
discussing: (i) a joint approach to solve all the abstraction sub-problems; (ii) a relaxation of the optimization problem; (iii) a parametrization of the relaxed problem; (iv) the enforcement of stochasticity on the parameters; (v) the enforcement of surjectivity on the parameters; (vi) solution by gradient descent.
We then present two immediate extensions of the algorithm: (vii) use of weighting on the loss function; and (viii) ensembling of models to better explore the solution space.

\paragraph{(i) Joint approach.} The abstraction learning problem consists of a set of related sub-problems. Simplistic approaches could ignore the structure connecting these problems. An \emph{independent approach}, for instance, would consider each problem separately, solve it, and, at the end, apply some form of aggregation to decide the value of those matrices appearing in multiple sub-problems. Similarly, a \emph{sequential approach} would establish an order among the sub-problems, and solve them accordingly, freezing the value of previously learned matrices. However, we aim at devising a \emph{joint approach} which, from the beginning, would take into considerations the existing structure and compute a jointly optimal solution over all the sub-problems at once. We will compare our joint approach against these simpler approaches and showcase the benefits of our algorithm.

\paragraph{(ii) Problem relaxation.}
Traversing the discrete solution space implied by a combinatorial optimization problem is notoriously challenging \citep{papadimitriou1998combinatorial} and the subject of current research \citep{titsias2017hamming,jaini2021sampling}.
%\theo{citations needed - also note that multiple mcmcm papers like ones i send can do that on principle}. 
We introduce a relaxation of the original problem, by redefining the original solution space $\mathcal{A}$ as
%\begin{equation}
    $\mathcal{A'} = \prod_{i=1}^{|\mathcal{X}|}[0,1]^{N_{i}\times M_{i}}$.
%\end{equation}
Solution matrices are now allowed to assume continuous values in $[0,1]$, which can be interpreted as allowing for stochastic or uncertain mappings. 
%Notice that, although the ground-truth abstraction matrices are expected to be binary by definition, relaxing this requirement may be interpreted as allowing for stochastic or uncertain mappings.\theo{can compress by taking info at end of last sentence and joining to first sentence. e.g. ``continuous values in $[0,1]$ that can be interpreted as allowing.."}

\paragraph{(iii) Parametrization of the problem.}
Having allowed for a continuous solution space, we now parametrize the problem by defining a set of parameters $\mathbf{W}$ containing a weight matrix $W_{X'} \in \mathbb{R}^{N_{i}\times M_{i}}$ for each abstraction $\alpha_{X'}$. We then redefine our learning problem as:
\begin{equation}\label{eqn:parametrizedProblem}
    \min_{\mathbf{W}\in\mathcal{A''}}e(\boldsymbol{\alpha}(\mathbf{W})),
\end{equation}
where the abstraction tuple $\boldsymbol{\alpha}(\mathbf{W}) = \langle R,a,\mathbf{W}\rangle$ is now parametrized by $\mathbf{W}$, and the solution space is
%\begin{equation}
    $\mathcal{A''} = \prod_{i=1}^{k}\mathbb{R}^{N_{i}\times M_{i}}$.
%\end{equation}

\paragraph{(iv) Enforcement of stochasticity.} 
The solution to the parametrized problem in Equation \ref{eqn:parametrizedProblem} returns a set of weights matrices in $\mathbb{R}^{N_{i}\times M_{i}}$. To force the solution towards a binary form, we apply a tempering operator column-wise
%\begin{equation}
    $\tmp{W}=\tfrac{e^{\frac{W_{i,j}}{T}}}{\sum_{i}e^{\frac{W_{i,j}}{T}}}$,
%\end{equation}
where $T \in \mathbb{R}_+$ is the temperature hyperparameter. This operator projects the solution into $[0,1]^{N_{i}\times M_{i}}$, approaching a binary solution in $\{0,1\}^{N_{i}\times M_{i}}$ as $T \rightarrow 0$. The problem can be re-expressed as:
\begin{equation} \label{eqn:firstlossterm}
\min_{\mathbf{W}\in\mathcal{A''}} \mathcal{L}_1 (\mathbf{W}) = \min_{\mathbf{W}\in\mathcal{A''}}  e(\boldsymbol{\alpha}(\tmp{\mathbf{W}})).
\end{equation}
Notice how the relaxation and the parametrization allowed us to move from the solution space $\mathcal{A}$ to $\mathcal{A'}$ to $\mathcal{A''}$, and how the tempering operator has allowed us to approximately project back to $\mathcal{A}$.

\paragraph{(v) Enforcement of surjectivity.} While the tempering operator returns a solution with the desired binary form, nothing guarantees that the solution matrices $\tmp{W_{X'}}$ will encode surjective functions. %Recall that, for a binary matrix $W_{X'}$ with shape $N_{i}\times M_{i}$ and $M_i\geq N_i$, encoding a surjective function means that every row contains at least a single one\theo{dont need to be so descriptive here - compress and just use math}. 
To enforce this property we introduce a second term in our loss function which penalizes every row in the learned matrices which do not contain at least a single one:
\begin{equation}\label{eqn:secondlossterm}
    \min_{\mathbf{W}\in\mathcal{A''}} \mathcal{L}_2 (\mathbf{W}) =  \min_{\mathbf{W}\in\mathcal{A''}} \sum_{W\in\mathbf{W}}\sum_{i=1}^{N_{i}}\left(1-\max_{j}\tmp{W}_{ij}\right).
\end{equation} 
%The amount of penalty is computed as follows: 
%For each matrix $W \in \mathbf{W}$, and for each row $i$, the penalty is given by the difference between 1 and the greatest element on the row.\theo{repeating info that exists on equation - compress or delete}
%; because of tempering, the greatest element in each row is bound in $[0,1]$.

\paragraph{(vi) Solution by gradient descent.}
Let our loss function be:
\begin{equation} \label{eqn:lossfunction}
	\min_{\mathbf{W}\in\mathcal{A''}} \mathcal{L}(\mathbf{W}) = 
	\min_{\mathbf{W}\in\mathcal{A''}} \lambda \mathcal{L}_1(\mathbf{W})  +  \mathcal{L}_2(\mathbf{W}),
\end{equation}
where $\lambda \in \mathbb{R}_{+}$ is a trade-off hyperparameter. $\mathcal{L}(\mathbf{W})$ is given by the sum of piecewise continuous functions: the first term is the supremum of JSDs, while the second term is related to the sum of maxima in the rows of the parameters. 
Given a random starting set of candidate solutions $\mathbf{W}$, it is possible to move in the solution space towards a locally optimal solution via gradient descent algorithms, iteratively computing $\mathbf{W} = \mathbf{W} - \eta \frac{\partial \mathcal{L}(\mathbf{W})}{\partial\mathbf{W}}$, with $\eta \in \mathbb{R}_+$ being a learning rate.

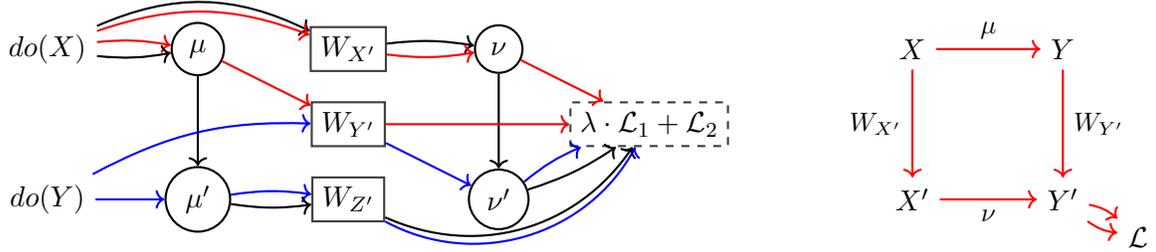
\begin{figure}
%\resizebox{\columnwidth}{!}{%
\centering
\begin{center}
\begin{tikzpicture}[node distance={15mm}, thick, main/.style = {draw, circle}] 
\node[dashed,thick,draw=black!75] (1) at (0,0) {$\lambda\cdot\mathcal{L}_1 + \mathcal{L}_2$}; 
\node[main] (2) at (-2,-1) {$\nu'$};
\node[rectangle,thick,draw=black!75] (3) at (-4,-1) {$W_{Z'}$};
\node[main] (4) at (-2,1) {$\nu$};
\node[rectangle,thick,draw=black!75] (5) at (-4,0) {$W_{Y'}$};
\node[rectangle,thick,draw=black!75] (6) at (-4,1) {$W_{X'}$};
\node[main] (7) at (-6,-1) {$\mu'$};
\node[main] (8) at (-6,1) {$\mu$};
\node[] (9) at (-8,-1) {$do(Y)$};
\node[] (10) at (-8,1) {$do(X)$};

\node[] (X) at (3.5,1) {$X$};
\node[] (Y) at (5.5,1) {$Y$};
\node[] (XX) at (3.5,-1) {$X'$};
\node[] (YY) at (5.5,-1) {$Y'$};

\node[] (L) at (6.5,-1.5) {$\mathcal{L}$};

\path[->,draw=blue] (2) edge [bend left=8] (1);
\path[->,draw=black] (2) edge [bend right=8] (1);
\draw[->,draw=red] (4) -- (1);
\draw[->,draw=black] (4) -- (2);
\path[->,draw=blue] (3) edge [bend right=45] (1);
\path[->,draw=black] (3) edge [bend right=40] (1);
\draw[->,draw=red] (5) -- (1);
\draw[->,draw=blue] (5) -- (2);
\path[->,draw=black] (6) edge [bend left=8] (4);
\path[->,draw=red] (6) edge [bend right=8] (4);
\path[->,draw=blue] (7) edge [bend left=8] (3);
\path[->,draw=black] (7) edge [bend right=8] (3);
\draw[->,draw=red] (8) -- (5);
\draw[->,draw=black] (8) -- (7);
\path[->,draw=black] (10) edge [bend left=25] (6);
\path[->,draw=red] (10) edge [bend left=20] (6);
\path[->,draw=red] (10) edge [bend left=8] (8);
\path[->,draw=black] (10) edge [bend right=8] (8);
\path[->,draw=blue] (9) edge (7);
\path[->,draw=blue] (9) edge [bend left=15] (5);

\path[->,draw=red] (X) to node[above,font=\small]{$\mu$} (Y);
\path[->,draw=red] (XX) to node[below,font=\small]{$\nu$} (YY);
\path[->,draw=red] (X) to node[left,font=\small]{$W_{X'}$} (XX);
\path[->,draw=red] (Y) to node[right,font=\small]{$W_{Y'}$} (YY);
\path[->,draw=red] (YY) edge [bend right=15] (L);
\path[->,draw=red] (YY) edge [bend left=15] (L);

\end{tikzpicture}
\end{center}

\caption{Neural network structure implied by the joint approach on an abstraction learning problem with $\mathcal{J} = \{ (X',Y'), (X',Z'), (Y',Z') \}$ (left). Circles represent known interventional distributions, solid rectangles learnable parameters, and the dashed box the loss function. Colors trace the diagrams defined by $(X',Y')$ (red), $(X',Z')$ (black), and $(Y',T')$ (blue); following, for instance, the red lines, it is possible reconstruct the upper and lower path as in the diagram of $(X',Y')$ (right).
%to reach the loss function via $\nu \circ W_{X'}$ (lower path) and $W_{Y'} \circ \mu$ (upper path). 
The network structure highlights the possibility of learning at once, via backpropagation, all the weight matrices shared by multiple diagrams. Contrast with other approaches in Appendix \ref{app:NN}.}
\label{fig:nn}
\end{figure}

Algorithm \ref{alg:jointlearning} in Appendix \ref{app:JointApproach} summarizes our joint approach. By relying on automatic differentiation we can see the algorithm as a form of neural network encoding the solution in the weights and having weight sharing defined by $\mathcal{J}$, as shown in Figure \ref{fig:nn}.

%\subsection{Extensions}

%We present two immediate extensions of the algorithm we have discussed: (i) use of collected interventional data for weighting our loss function; (ii) ensembling of models to better explore the solution space.

\paragraph{(vii) Weighting by interventional data.}
The loss term in Equation \ref{eqn:firstlossterm} implicitly weights each abstraction error $E_{\abs}(\mathbf{X'},\mathbf{Y'})$ uniformly. It may be desirable, however, to scale the error with the relevance of the different interventional distributions using a vector $\boldsymbol{\kappa} \in \mathbb{R}_{+}^{|\mathcal{J}|}$:
\begin{eqnarray}
	&\min_{\mathbf{W}\in\mathcal{A}''}&\mathcal{L}_{1}(\mathbf{W},\boldsymbol{\kappa})=\min_{\mathbf{W}\in\mathcal{A}''}e(\boldsymbol{\alpha}(\tmp{\mathbf{W}},\boldsymbol{\kappa}))  =\\
	&\min_{\mathbf{W}\in\mathcal{A}''}&\sup\left\{ \kappa_{1}D_{JSD}(\alpha_{\mathbf{Y'}}\mu;\nu\alpha_{\mathbf{X'}}),\kappa_{2}D_{JSD}(\alpha_{\mathbf{Z'}}\mu';\nu'\alpha_{\mathbf{Y'}})...\right\}.
\end{eqnarray}
Assuming that more relevant interventions are collected more often, these weights $\boldsymbol{\kappa}$ may be set proportionally to the amount of interventions collected.  Alternatively, importance schemes re-weighting samples or interventions \citep{xu2021understanding} or affecting the balance between JSD losses \citep{vandenhende2021multi} may be adopted.

\paragraph{(viii) Ensembling of models.}
Solving a relaxed combinatorial optimization problem by gradient descent does guarantee only the achievement of a local optimum \citep{papadimitriou1998combinatorial}. In particular, gradient descent is sensitive to the morphology of the loss landscape and parameter initialization. Ensembling has been shown to improve the performance and the uncertainty estimation of learning \citep{dietterich2000ensemble,lakshminarayanan2017simple}; we then rely on running an ensemble of models with different starting parameters $\mathbf{W}$ in order to learn better abstractions.%\theo{This is pretty obvious and well-known - not much info in this paragraph. Perhaps point to literature where ensemble approaches have been shown to work well in achieving a better overall solution - e.g. deep learning}

\section{Experiments} \label{sec:Experiments}

In this section we report results from running our algorithm both on synthetic and real-world data. Data and code for all simulations are openly available at \url{https://github.com/mattdravucz/jointly-learning-causal-abstraction/}.

\subsection{Synthetic experiments}
In our synthetic simulations, we consider four scenarios featuring different aspects of abstraction and presenting different challenges. 
%(i) the \emph{basic lung cancer} scenario presented in \cite{rischel2020category} where we are given chain models, and abstraction consists in the exclusion of a low-level variable; (ii) a \emph{collapsing lung cancer} scenario where we have chain models, and abstraction consists of the merging of low-level variables; (iii) an \emph{extended lung cancer} scenario where we have chain models and abstraction consists in the reduction of the resolution of the endogenous variables; (iv) a \emph{v-structure lung cancer} scenario where we have v-structures and abstraction consists in the elimination and collapsing of variables as well as in the reduction of resolution. 
Table \ref{tab:SyntheticExperiments} provides an overview of these scenarios; as evinced from it, our scenarios cover different forms of abstractions, while keeping the complexity of the problems limited for verification via enumeration. Appendix \ref{app:SyntheticData_models} provides details for all the models, the abstractions, the set $\mathcal{J}$, and the optimal solutions computed via enumeration.

\begin{table}
\caption{Summary of the scenarios. \emph{Scenario} and \emph{Type of abstraction} describe the scenario; \emph{Sub-pr.} refers to the number of sub-problems (and hence the number of diagrams) implied by each abstraction learning problem; \emph{Abstraction Matrices} lists the shapes of the abstraction matrices to be learned; \emph{\#Sol.} evaluates the number of possible solutions according to Equation \ref{eqn:all_n_surj_functions}; \emph{Optimum} reports the existence of a unique zero-error or non-zero error solution. \label{tab:SyntheticExperiments}}
\resizebox{\columnwidth}{!}{%
\begin{tabular}{c>{\centering}p{4.5cm}cccc}
\hline 
\textbf{\emph{Scenario}} & \textbf{\emph{Type of abstraction}} & \textbf{\emph{Sub-pr.}} & \textbf{\emph{Abstraction Matrices}} & \textbf{\emph{\#Sol.}} & \textbf{\emph{Optimum}}\tabularnewline
\hline 
Basic & Elimination of low-level var & 1 & $\{2\times2,2\times2\}$ & 4 & Unique zero-error\tabularnewline
\hline 
Collapsing & Merging of low-level vars & 3 & $\{2\times4,2\times2,2\times2\}$ & 56 & Unique non-zero-error\tabularnewline
\hline 
Extended & Reduction of resolution & 3 & $\{3\times4,2\times3,2\times2\}$ & 432 & Unique zero-error\tabularnewline
\hline 
v-Structure & Elimination of low-level var\\
Merging of low-level vars\\
Reduction of resolution & 3 & $\{2\times2,2\times2,2\times4\}$ & 56 & Unique non-zero-error\tabularnewline
\hline 
\end{tabular}}
\end{table}

For each one of these scenarios, we first compute the ground-truth solution via enumeration. Then we perform the following studies: (a) we run our algorithm, and compare its performance against an independent and a sequential baseline approach (see Section \ref{sec:Methodology}(i)); (b) we analyze how weighting can affect the quality of the solutions; (c) we perform an ablation study in which we remove the surjectivity penalty term; (d) we perform a sensitivity analysis in which we vary the value of the hyperparameters $T$ and $\lambda$ specified by our approach.
Performance is evaluated in terms of JSD, surjective penalty, $\lone$-distance from the optimal ground-truth solution, and wall-clock time. Results are averaged over 10 simulations. Algorithms for all the approaches are provided in Appendix \ref{app:SolutionApproaches} and details about the experimental settings and metrics in Appendix \ref{app:SyntheticData_experimental}. 

Figure \ref{fig:syntexp_a_performance} exemplifies the learning process in the \emph{extended} scenario. The joint approach reliably learns a solution closer to the ground-truth optimum than the other approaches. As low levels of JSD and surjective penalty are necessary but not sufficient to reach the optimal ground-truth solution, the independent and sequential approach achieve a low loss, but their $\lone$-distance is significantly higher than the joint approach. We hypothesize that the better results of our algorithm are due to the reliance on the information shared between sub-problems that is discarded by the other approaches. Analogous plots for the other scenarios are available in Appendix \ref{app:SyntheticData_additional}. 
Table \ref{tab:syntexp_a_performance} provides the performance of the three approaches aggregated over the four scenarios. Consistently with our hypothesis, the joint approach performs better or equally to the other algorithms in terms of $\lone$-distance from the ground truth by exploiting all the shared information; next comes the sequential approach which uses shared information only partially; last is the independent approach which completely discards it. For an analogous reason, ordering in terms of time efficiency is reversed: dropping shared information, the independent approach is fully parallelizable wrt the sub-problems; the sequential approach may deem some sub-problems redundant and avoid solving them; the joint approach considers all the sub-problems jointly at once.
Table \ref{tab:syntexp_b_performance} shows how the final result of the joint algorithm may be affected by weighting. In this instance, we considered only the \emph{v-structure} scenario, and we assumed that the JSD error related to cancer when intervening on smoking ($E_{\abs}(S',C')$) would be more important than the JSD error for fatigue when intervening on cancer ($E_{\abs}(C',F')$) or when intervening on smoking ($E_{\abs}(S',F')$). When placing $80\%$ of the weight on $E_{\abs}(S',C')$, the final JSD for this interventional diagram decreases, while other JSD do not significantly change and exhibit higher variance. Weighting can then be used to get better approximations on those parts of the problem the modeller is more concerned with.
Table \ref{tab:syntexp_c_performance} confirms the the critical role of the surjective penalty $\mathcal{L}_2$; in its absence the algorithm can learn a better solution in terms of JSD by ignoring values in the abstracted model, but it lands on a solution significantly further from the optimal solution. Additional discussion and sample learned matrices are provided in Appendix \ref{app:SyntheticData_additional}, together with results from the sensitivity analysis.

\begin{figure}
%\resizebox{\columnwidth}{!}{%
\centering
\includegraphics[scale=0.5]{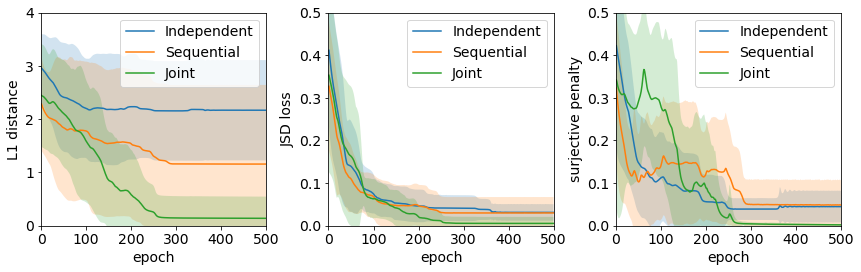}
%\resizebox{1.0\textwidth}{!}{\input{img/apgf-1test.pgf}}
\caption{Performance during training on the \emph{extended} scenario. While all approaches minimize their loss, the joint approach achieves a significantly lower (Wilcoxon test, $p\textrm{-value}<0.05$) $\lone$-distance from the ground truth.}
\label{fig:syntexp_a_performance}
\end{figure}

\begin{table}
\caption{Performance at the end of training aggregated over the four scenarios. In all instances, the joint approach competes or overcomes the baseline approaches, although with longer runtime.}
\label{tab:syntexp_a_performance}
\centering
\resizebox{.7\columnwidth}{!}{%
\begin{tabular}{ccccc}
\hline 
 & \textbf{\emph{L1 Dist}} & \textbf{\emph{JSD Loss}} & \textbf{\emph{Surj Pen}} &  \textbf{\emph{Time}}\tabularnewline
\hline 
Independent & $3.25\pm2.06$ & $0.50\pm0.31$ & $0.40\pm0.37$ & ${3.74\pm0.05}$ \tabularnewline
\hline 
Sequential &  $1.45\pm1.55$ & $0.28\pm0.09$ & ${0.03\pm0.07}$ & $7.10\pm0.41$ \tabularnewline
\hline 
Joint &  ${0.85\pm0.64}$ & ${0.24\pm0.03}$ & $0.08\pm0.11$ & $8.12\pm0.20$ \tabularnewline
\hline 
\end{tabular}}
\end{table}

\begin{table}
\centering
\begin{minipage}[t]{0.53\linewidth}\centering
\caption{JSD on the \emph{v-structure} scenario. Weighting causes a marginal redistribution of errors.}
\label{tab:syntexp_b_performance}
\resizebox{\columnwidth}{!}{%
\begin{tabular}{cccc}
\hline 
 & $E_{\boldsymbol{\alpha}}(S',C')$ & $E_{\boldsymbol{\alpha}}(C',F')$ & $E_{\boldsymbol{\alpha}}(S',F')$\tabularnewline
\hline 
Unweighted & $0.28\pm0.07$ & $0.28\pm0.01$ & ${0.15\pm0.03}$ \tabularnewline
\hline 
Weighted &  ${0.25\pm0.05}$ & ${0.27\pm0.06}$ & ${0.15\pm0.04}$ \tabularnewline
\hline 
\end{tabular}}
\end{minipage}\hfill%
\begin{minipage}[t]{0.43\linewidth}\centering
\caption{Performance on the \emph{v-structure} scenario. Surjectivity penalty is critical to achieve significantly better (Wilcoxon test, $p\textrm{-value}<0.05$) results.}
\label{tab:syntexp_c_performance}
\resizebox{\columnwidth}{!}{%
\begin{tabular}{ccc}
\hline 
 & \emph{\textbf{L1 Dist}} & \emph{\textbf{JSD Loss}} \tabularnewline
\hline 
Joint & $\mathbf{3.00\pm1.34}$ & $0.72\pm0.08$  \tabularnewline
\hline 
Ablated joint & $5.00\pm2.41$ & ${0.68\pm0.15}$  \tabularnewline
\hline 
\end{tabular}}
\end{minipage}
\end{table}

\subsection{Real-world experiment}

We learn an abstraction between the implicit causal models for lithium-ion battery manufacturing developed by two research groups: the Laboratoire de Réactivité et Chimie des Solides (LRCS) group, and the Warwick Manufacturing Group (WMG). 
%Specifically, we focus on the manufacturing step of cathode material \emph{battery coating} \citep{cunha2020artificial,liu2022interpretable}, and how the main control variable \emph{Comma Gap} (CG) affect the output variables \emph{Mass Loading} (ML). 
Battery electrode manufacturing is a complex process involving several key stages (e.g. material selection, mixing and coating.%, calendaring, cutting, electrolyte filling and testing). 
To develop high-performing batteries, it is necessary to understand how each of the manufacturing parameters influence the subsequent product. The current approach relies heavily on experienced lab personnel with extensive knowledge in order to adjust the manufacturing parameters and achieve a desired battery performance. Considerable research effort is directed to develop models of the manufacturing stages \citep{RomanRamirez2022b,Niri2022} and optimised feedback control mechanisms to reduce the reliance on human expertise.

In this work we focus on the dry mass loading ($ML$) variable from the coating stage \citep{cunha2020artificial,liu2022interpretable}, which directly determines the energy density of the final battery. The dry ML is in part controlled by the comma-bar gap ($CG$) variable, which is set manually and guides the mass loading of the active material in its wet form; the wet coating subsequently passes through a drying stage resulting in the dry coating. Public datasets on this process are scarce and of limited dimensions, due to the cost and complexity of the measurements; however, the ability to predict the dry ML based on the upstream CG variable is vital in order to achieve a target battery energy density and increase manufacturing efficiency. To obviate this problem, we aim at learning an abstraction that may relate the models assumed by the two research groups and then integrate their data to significantly improve downstream inferences.

We use a dataset for battery coating from the LRCS group \citep{cunha2020artificial}, and recordings performed by the WMG group. We perform pre-processing in order to select the relevant variables. 
As in many real-world scenarios, fully-specified SCMs are not available, so we define elementary SCMs $\mathcal{M}^{WMG}$ and $\mathcal{M}^{LRCS}$ with minimal assumptions (Figure \ref{fig:LRCR_SCM} and \ref{fig:SCM_WMG}).
%while defining nodes and edges is in our case is obvious, specifying the mechanisms is not trivial, and we rely on the available data to learn them. Once we have setup the two SCMs, $\mathcal{M}^{WMG}$ and $\mathcal{M}^{LRCS}$, we decide to learn an abstraction from $\mathcal{M}^{WMG}$ to $\mathcal{M}^{LRCS}$.
We then elect the WMG model as the base model since it has a higher resolution in terms of number and domain of observed variables. Appendix \ref{app:RealData_models} provides details about the data, pre-processing, the SCMs and the abstraction.
We learn the abstraction $\abs{}$ from $\mathcal{M}^{WMG}$ to $\mathcal{M}^{LRCS}$ using our joint approach, similarly to what we have done with the synthetic experiments. We evaluate results in two ways. Qualitatively, we assess the shape of the learned matrices, to confirm they are binary and surjective, and to check whether identical values in the domain of the base and abstracted variables are mapped to each other. Quantitatively, we set three downstream regression problems, see Table \ref{tab:realexp_performance}, to assess whether transporting data via abstraction could help improve prediction. Task (a) represents a challenging scenario in which we use limited and expensive experimental data to perform predictions. Task (b) represents a scenario in which data transported via abstraction provides support for our predictions, while task (c) the case in which transported data augments the training set but does not provide support on the test set. See Appendix \ref{app:RealData_experimental} for details on experimental settings and metrics.

Figure \ref{fig:realexp_mat_cg} shows the learned matrix $W_{CG}$ corresponding to the abstraction map $\alpha_{CG}$ relating the $CG$ variables in the two models. The matrix approximates a binary matrix encoding a surjective function. A red border is used to denote identical value of the $CG$ matrix in the base and abstracted model; these values are successfully mapped to each other, while intermediate values align along the main diagonal.
%the values that would map $CG=75$ in $\mathcal{M}^{WMG}$ to $CG=75$ in $\mathcal{M}^{LRCS}$, and similarly $CG=200$ in $\mathcal{M}^{WMG}$ to $CG=200$ in $\mathcal{M}^{LRCS}$; intermediate values would be expected along the diagonal. 
Notice, however, that alternative patterns could also be learned, as discussed in Appendix \ref{app:RealData_additional}, together with shape of the learned matrix $W_{ML}$.

Table \ref{tab:realexp_performance} shows the mean-square error when learning only on the small LRCS dataset (task a) and when using data transported via abstraction from the WMG dataset; performance improves both when WMG data provides the missing support for prediction (task b) or when it just augments the dataset (task c).; when not providing the required missing support, we observe an improvement on selected cases; this is likely due to having learned a non-perfect abstraction and to the noise introduced during the transport of the data. %\theo{You wanna finish stronger here rather then on a weak point. Change order of things or say something at the end on the positive side}

\begin{table}
\begin{minipage}[t]{0.33\linewidth}\centering
\resizebox{\columnwidth}{!}{%
\includegraphics{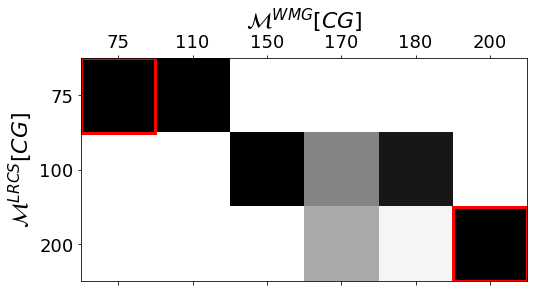}}
\captionof{figure}{Learned $W_{CG}$. The matrix is close to binary, surjective, and diagonal as expected.}
\label{fig:realexp_mat_cg}
\end{minipage}\hfill%
\begin{minipage}[b]{0.63\linewidth}\centering
\caption{Regression problems setup and MSE. Transporting data via abstraction reduces the error.}
\label{tab:realexp_performance}
\resizebox{\columnwidth}{!}{%
\begin{tabular}{cccc}
\hline 
 & \textbf{\emph{Training set}} & \textbf{\emph{Test Set}} & \textbf{\emph{MSE}} \tabularnewline
\hline 
(a) & LRCS{[}$CG\neq k${]} & LRCS{[}$CG=k${]} & $1.86\pm1.75$\tabularnewline
\hline 
(b) & LRCS{[}$CG\neq k${]} & LRCS{[}$CG=k${]} & $0.22\pm0.26$ \tabularnewline
& + WMG &   & \tabularnewline
\hline 
(c) & LRCS{[}$CG\neq k${]} & LRCS{[}$CG=k${]} & $1.22\pm0.95$ \tabularnewline
& + WMG{[}$CG\neq k${]} & + WMG{[}$CG=k${]} & \tabularnewline
\hline 
\end{tabular}}

\end{minipage}
\end{table}

\section{Discussion and Conclusion} \label{sec:Discussion}
%\theo{too much stuff here - compress as the real info is not that much. For example first paragraph rephrases and refines sequentially the same point 3 times. Be more spartan here we need half the size of this section.}

In this paper we have built an abstraction learning framework starting from the formalization of an abstraction between SCMs proposed in \citep{rischel2020category}. We have examined how we could use this definition to express the learning problem as an abstraction error minimization problem, and proposed an algorithm to solve it. Our approach has been based on relaxing the initial problem, parametrizing it and solving it via gradient descent. Results on synthetic datasets show that our algorithm perform better than other simpler approaches, and it reaches a solution closer to the optimum. Furthermore, application to EV manufacturing data provided a proof of concept of the usefulness of learning abstraction to relate models and transport data in low-data regimes.

The abstraction learning problem we have tackled is a particular instance of the very generic problem of learning a mapping not just between isolated sets or objects, but between sets and objects sharing a structure expressed by transformations or morphisms. 
%Although we worked on a problem defined over specific mathematical models (in a given category), the underlying problem has a very generic form that expresses the challenge of finding mappings between sets (or concepts) that would be consistent under certain transformations (or mechanisms); that is, a problem where we learn not only a mapping between isolated sets (or objects), but where we learn a mapping that preserves a structure defined by relevant transformations (or morphisms).
This problem has been given the form of a notoriously challenging discrete combinatorial optimization problem, for which other common heuristics exist in the optimization literature. 
%Our joint approach solves a collection of interconnected discrete optimization sub-problems via differentiable programming. 
Application of our approach to the real-world data of a lithium-ion battery manufacturing stage provided a proof of concept of the potential of learning abstraction between models devised by different groups, although more research would be needed to make the proposed algorithm widely usable in real-world applications. The joint approach is sensitive to initialization, and using abstraction to transfer data may introduce noise; increasing robustness and extending it to the continuous domain are directions for future work. %However, other common heuristics in optimization may be explored to solve the abstraction learning problem.

As a first instance of an abstraction learning problem, our setup assumes perfect knowledge of the models. Dropping this requirement leads to problems in which we could learn both abstraction and distributions from data (incidentally, in our real-world scenario, we trivially learned mechanisms from data at setup time, but the learning of mechanisms and abstraction could happen jointly).
Similarly, we may further limit knowledge about the abstraction, dropping, for instance, the requirement of knowing $a$ or $R$ \citep{zennaro2022towards}.
%; this leads to more challenging abstraction learning problems, as presented in the hierarchy in \cite{zennaro2022towards}. 
Alternatively, it may also be possible to consider using domain knowledge on how specific variables and outputs are related to guide learning; the space of surjective function grows very rapidly, and restricting it by using available priors would simplify the learning problem.
Other relevant directions of work include theoretical evaluation of our relaxation or the definition of the set $\mathcal{J}$
%which determines the interventional distributions to be considered. 
(as discussed in Appendix \ref{app:J}, $\mathcal{J}$ may contain irrelevant or redundant interventional distributions, and an algorithm that selects relevant interventions may take advantage of ideas such as minimal intervention sets \citep{lee2018structural}).

\acks{TD acknowledges support from a UKRI Turing AI acceleration Fellowship [EP/V02678X/1]. The WMG data was undertaken as part of the NEXTRODE project funded by The Faraday Institution, UK [Grant Number: FIRG015]. The authors would also like to thank Dr Michael Lain at WMG for supporting with collecting battery coating data.}

\bibliography{citation}

\newpage
\appendix

\section{Notation} \label{app:Notation}

\resizebox{\columnwidth}{!}{%
\begin{tabular}{cc}
\hline 
$\mathcal{M},\mathcal{M}'$ & SCMs\tabularnewline
$\mathcal{X},\mathcal{X}'$ & Set of endogenous variables (for $\mathcal{M}$ and $\mathcal{M}'$)\tabularnewline
$\mathcal{U},\mathcal{U}'$ & Set of exogenous variables (for $\mathcal{M}$ and $\mathcal{M}'$)\tabularnewline
$\mathcal{F},\mathcal{F}'$ & Set of structural functions (for $\mathcal{M}$ and $\mathcal{M}'$)\tabularnewline
$P(\mathcal{U})$,$P'(\mathcal{U}')$ & Joint distributions for exogenous variables (in $\mathcal{M}$ and
$\mathcal{M}'$)\tabularnewline
$X,Y,Z,X_{i},...,X',Y',Z',X_{i}',...$ & Endogenous variables (in $\mathcal{M}$ and $\mathcal{M}'$)\tabularnewline
$\mathbf{X},\mathbf{Y}...\mathbf{X'},\mathbf{Y}'...$ & Sets of endogenous variables (in $\mathcal{M}$ and $\mathcal{M}'$)\tabularnewline
$\mathcal{M}[X],\mathcal{M}'[X']$ & Domain of endogenous variable (in $\mathcal{M}$ and $\mathcal{M}'$)\tabularnewline
$f_{1},f_{2},...,f'_{1},f'_{2},...$ & Structural functions (in $\mathcal{M}$ and $\mathcal{M}'$)\tabularnewline
$\mathcal{G}_{\mathcal{M}}$ & DAG underlying the SCM $\mathcal{M}$\tabularnewline
$pa(X)$ & Parents of node $X$\tabularnewline
\hline 
$\iota \coloneqq do(X_{i}=x_{i})$ & Intervention\tabularnewline
$\mathcal{M}_{\iota},\mathcal{M}_{\iota'}'$ & Post-interventional SCMs\tabularnewline
\hline 
$\boldsymbol{\alpha}$ & Abstraction tuple\tabularnewline
$R$ & Set of relevant nodes\tabularnewline
$a$ & Structural-level surjective map\tabularnewline
$\alpha$ & Collection of distributional-level surjective maps\tabularnewline
$\alpha_{X'},\alpha_{Y'},...$ & Distributional-level surjective maps encoded as binary stochastic
matrices\tabularnewline
$\mu,\nu,\mu',\nu'...$ & Interventional distributions encoded as stochastic matrices\tabularnewline
$\mathcal{J}$ & Set of pair sets of endogenous variables or set of interventional
distributions\tabularnewline
$E_{\boldsymbol{\alpha}}(\mathbf{X',Y'})$ & Abstraction error wrt pair set $(\mathbf{X',Y'})$ or wrt interventional
 $P'(\mathbf{Y}'\vert do(\mathbf{X}'))$\tabularnewline
$e(\boldsymbol{\alpha})$ & Overall abstraction error for $\boldsymbol{\alpha}$\tabularnewline
\hline
\end{tabular}}

\section{Algebraic encoding of a SCM}\label{app:AlgebraicEncoding}

We illustrate here how a SCM $\mathcal{M}$ defined on a finite set of variables with finite domains may be expressed as a collection of sets and stochastic matrices. A more formal treatment is given by \cite{rischel2020category}.

Let $\mathcal{M} = \langle \mathcal{X},\mathcal{U},\mathcal{F},P(\mathcal{U}) \rangle$
be a SCM with underlying DAG $\mathcal{G}_{\mathcal{M}}$.

Each endogenous variable $X_i \in \mathcal{X}$ can be immediately associated with its domain, that is, the discrete set: $$\mathcal{M}[X_i]=\{0,1,...,M\}.$$
When considering a subset of variables $\mathbf{X} = \{X_1, X_2, ..., X_k\} \subseteq \mathcal{X}$, the associated set is the Cartesian product of the domain of each variable, that is, $\mathcal{M}[\mathbf{X}]=\mathcal{M}[X_1] \times \mathcal{M}[X_2] \times ... \times \mathcal{M}[X_k]$.

Thanks to the measurability of the structural functions in $\mathcal{F}$, the probability distribution $P(\mathcal{U})$ over the exogenous variables can be pushforwarded over the endogenous variables; this allows us to define a joint distribution $P(\mathcal{X})$ over the endogenous variables \citep{peters2017elements}. The joint distribution can then be factored according to the structure of the DAG, allowing us to compute distributions $P(X_i)$ on root nodes and conditional distributions $P(Y_i\vert pa(Y_i))$ on non-root nodes. A distribution $P(X_i)$ on a $M$-dimensional set $\mathcal{M}[X_i]=\{0,1,...,M\}$ can be represented as a stochastic $M \times 1$ matrix:
$$
\left[\begin{array}{c}
p_{1}\\
p_{2}\\
...\\
p_{M}
\end{array}\right],
$$
with the constraint that $\sum_{j=1}^{M} p_j = 1$.
A distribution $P(Y_i \vert pa(Y_i))$, where the set associated with $pa(Y_i)$ is $M$-dimensional set and the set associate with $Y_i$ is $N$-dimensional, can be represented as a stochastic $N \times M$ matrix:
$$
\left[\begin{array}{cccc}
p_{1,1} & p_{1,2} & ... & p_{1,M}\\
p_{2,1} & p_{2,2} & ... & p_{2,M}\\
... & ... & ... & ...\\
p_{N,1} & p_{N,2} & ... & p_{N,M}
\end{array}\right],
$$
with the constraint that, for each $0\leq k \leq M$, $\sum_{j=1}^{N} p_{j,k} = 1$. This matrix can be seen as encoding in each column $k$ a single conditional distribution $P(Y_i \vert pa(X_i))$ for each of the $M$ values that the conditioning variable $pa(X_i)$ can assume. Notice that, from the joint distribution, we can also derive in the same way a matrix for any other marginal or conditional distribution not necessarily associated with the factorization defined by the DAG.

It is worth noting that this reduction of a SCM to a collection of sets and stochastic matrices does not violate Pearl's hierarchy \citep{bareinboim2022pearl} as it does not claim an equivalence between a SCM and its algebraic reduction; indeed, after algebraic reduction, computing interventions and counterfactuals is not possible anymore.

Finally, an abstraction map $\alpha_{X'}$ is a surjective function from a (set of) low-level variable(s) with cardinality $M$ to a high-level variable with cardinality $N$. Notice that, for surjectivity to be possible, it must hold $M \geq N$. This matrix can again be represented as a stochastic matrix:
$$
\left[\begin{array}{cccc}
\alpha_{1,1} & \alpha_{1,2} & ... & \alpha_{1,M}\\
\alpha_{2,1} & \alpha_{2,2} & ... & \alpha_{2,M}\\
... & ... & ... & ...\\
\alpha_{N,1} & \alpha_{N,2} & ... & \alpha_{N,M}
\end{array}\right],
$$
with two constraints: (i) functionality: every column $k$ contains a single one and $N-1$ zeros, thus encoding a degenerate deterministic distribution; (ii) surjectivity: every row $j$ contains at least a one.

Through this encoding, all the quantities in the diagram in Definition \ref{def:AbstractionConsistency} can be expressed algebraically.

\section{Definition of Jensen-Shannon distance} \label{app:JSD}

We recall here the definition of the discrete Kullback–Leibler divergence the discrete Jensen-Shannon distance \citep{cover1999elements}. 

\begin{definition}[Kullback–Leibler (KL) divergence]
    Let $p$ and $q$ be two probability mass functions on the same domain $\mathcal{X}$, with $q(x)>0$  $\forall x\in\mathcal{X}$. The KL divergence from $p$ to $q$ is defined as:
    \begin{equation}
        D_{KL}(p;q) = -\sum_{x\in\mathcal{X}} p(x) \log \frac{p(x)}{q(x)}.
    \end{equation}
\end{definition}

\begin{definition}[Jensen-Shannon (JSD) distance]
    Let $p$ and $q$ be two probability mass functions on the same domain $\mathcal{X}$, with $p(x)>0$ and $q(x)>0$ $\forall x\in\mathcal{X}$. The JS distance between $p$ and $q$ is defined as:
    \begin{equation}
        D_{JSD}(p;q) = \sqrt{\frac{1}{2} D_{KL}(p;m) + \frac{1}{2} D_{KL}(q;m)},
    \end{equation}
    where $m = \frac{1}{2} p + \frac{1}{2} q$.
\end{definition}

\section{Choice of $\mathcal{J}$} \label{app:J}

The set $\mathcal{J}$ provides the list of interventional distributions (i.e., diagrams or sub-problems) to be considered in the abstraction learning problem.

According to Definition \ref{def:OverallAbstractionError}, $\mathcal{J}$ contains all disjoint pair sets $(\mathbf{X'},\mathbf{Y')} \in \mathscr{P}(\mathcal{X'})\times\mathscr{P}(\mathcal{X'})$; such a set $\mathcal{J}$ may be computed by taking the Cartesian product $\mathscr{P}(\mathcal{X'}) \times \mathscr{P}(\mathcal{X'})$ of the power sets and by selecting only those elements such that $\mathbf{X'} \cap \mathbf{Y'} = \emptyset$. However, this set $\mathcal{J}$, beyond being very large, might contain many pairs corresponding to irrelevant or redundant interventional distributions. 

Irrelevant interventional distributions correspond, for instance, to anti-causal interventions; given a DAG where we have $X' \rightarrow Y'$, the set $\mathcal{J}$ will contain the set $(X',Y')$ corresponding to the meaningful interventional distribution $P'(Y'\vert do(X'))$, but also the set $(Y',X')$ corresponding to the anti-causal interventional distribution $P'(X' \vert do(Y'))$). Redundant interventional distributions correspond to interventions on independent variables; for instance,  given a DAG where we have $X' \rightarrow Y' \rightarrow Z'$, the set $\mathcal{J}$ will contain the set $(Y',Z')$ corresponding to the minimal interventional distribution $P'(Z'\vert do(Y'))$, but also the set $((X',Y'),Z')$ corresponding to the redundant interventional distribution $P'(Z' \vert do(X',Y'))$), where $Z'$ is independent from $X'$ once we intervene on $Y'$.

Devising a proper algorithm that select a minimal set $\mathcal{J}$ of relevant interventional distributions is left to future work, and discussed in Section \ref{sec:Discussion}. In our simulations, we manually select meaningful interventional distributions wrt to the application at hand.

\section{Details about the solution approaches} \label{app:SolutionApproaches}

We provide the pseudocode for our joint approach and for the approaches discussed in Section \ref{sec:Methodology}(i) and used in Section \ref{sec:Experiments}.

\subsection{Joint approach} \label{app:JointApproach}

Algorithm \ref{alg:jointlearning} presents the pseudocode of our joint learning approach.

\begin{algorithm}
	\caption{Joint Learning Algorithm}\label{alg:jointlearning}
	\begin{algorithmic}[1]
		\Require base model $\mathcal{M}$, abstracted model $\mathcal{M'}$, set of relevant variables $R$, mapping $a$, temperature $T$, trade-off $\lambda$, learning rate $\eta$, $n_e$ number of learning epochs
		\Ensure locally optimal collection $\alpha$ of abstraction maps $\alpha_{X'}$
		
		\State Instantiate the parameter set $\mathbf{W} = \{\}$ \Comment{Setup the params}
		\State \For{$X' \in \mathcal{X'}$ }{		
			Instantiate a matrix $W_{X'}$ with dimension $|\mathcal{M'}[X']| \times |\mathcal{M}[a^{-1}(X')]|$
			
			Add matrix $W_{X'}$ to $\mathbf{W}$
		} 
		
		\State Instantiate the loss $\mathcal{L}_1 = \sup \{ \}$	\Comment{Setup the first term in the loss}
		\State Retrieve the set $\mathcal{J}$
		\State \For{$j = (X',Y') \in \mathcal{J}$}{
			Add the term $D_{JSD}(\tmp{W_{X'}}\mu; \nu \tmp{W_{Y'}})$ to $\mathcal{L}_1$  
		}
		
		\State Instantiate the loss $\mathcal{L}_2 = \sum \{ \}$	\Comment{Setup the second term in the loss}
		\State \For{$W \in \mathbf{W}$ }{		
			Add $\sum_{i}\left(1-\max_{j}\tmp{W}_{ij}\right)$ to $\mathcal{L}_2$
		} 
		
		\State Assemble the loss $\mathcal{L} = \lambda \mathcal{L}_1 +  \mathcal{L}_2$ \Comment{Setup the loss}
		\State \For{$n_e$}{
			Optimize by gradient descent $\mathbf{W} = \mathbf{W} - \eta \frac{\partial \mathcal{L}(\mathbf{W})}{\partial\mathbf{W}}$ \Comment{Learn}
		}
	\end{algorithmic}
\end{algorithm}

The choice of the set $\mathcal{J}$ at step 4 is discussed in Appendix \ref{app:J}. Also, notice that setting up the first term of the loss in the loop at step 5 is the core of the algorithm, where knowledge about the models and the abstraction is exploited.

\subsection{Independent approach} \label{app:IndependentApproach}

Algorithm \ref{alg:indeplearning} presents the pseudocode of the independent learning approach.

\begin{algorithm}
	\caption{Independent Learning Algorithm}\label{alg:indeplearning}
	\begin{algorithmic}[1]
		\Require base model $\mathcal{M}$, abstracted model $\mathcal{M'}$, set of relevant variables $R$, mapping $a$, temperature $T$, trade-off $\lambda$, learning rate $\eta$, $n_e$ number of learning epochs
		\Ensure locally optimal collection $\alpha$ of abstraction maps $\alpha_{X'}$
		
		\State Instantiate the parameter set $\mathbf{W} = \{\}$
		\State Retrieve the set $\mathcal{J}$
		\State \For{$j = (X',Y') \in \mathcal{J}$}{
			Setup $\mathbf{\tilde{W}} = \{\}$  \Comment{Setup the params}
			
			Instantiate matrices $W_{X'}$ and $W_{Y'}$ and add them to $\mathbf{\tilde{W}}$ 
			
			Instantiate $\mathcal{L}_1 = D_{JSD}(\tmp{W_{X'}}\mu; \nu \tmp{W_{Y'}})$ \Comment{Setup the first term in the loss} 
			
			Instantiate $\mathcal{L}_2 = \sum_{i}\left(1-\max_{j}\tmp{W_{X'}}_{ij}\right) + \sum_{i}\left(1-\max_{j}\tmp{W_{Y'}}_{ij}\right)$ \Comment{Setup the second term in the loss} 
			
			Assemble the loss $\mathcal{L} = \lambda \mathcal{L}_1 + \mathcal{L}_2$ \Comment{Setup the loss}
			
			\For{$n_e$}{
				Optimize by gradient descent $\mathbf{\tilde{W}} = \mathbf{\tilde{W}} - \eta \frac{\partial \mathcal{L}(\mathbf{\tilde{W}})}{\partial(\mathbf{\tilde{W}})}$ \Comment{Learn}
			}
			
			Add solution $\mathbf{\tilde{W}}$ to $\mathbf{W}$
		}
		
		\State\For{repeated $W_{X'} \in \mathbf{W}$}{Aggregate by majority voting \Comment{Aggregation}}
	\end{algorithmic}
\end{algorithm}

The independent learning approach follows an algorithm similar to the joint learning algorithm; however, it solves an optimization problem for each sub-problem in $\mathcal{J}$. Notice that the majority voting algorithm in step 4 may be substituted by any other aggregation algorithm.

\subsection{Sequential approach} \label{app:SequentialApproach}

Algorithm \ref{alg:seqlearning} presents the pseudocode of the sequential learning approach.

\begin{algorithm}
	\caption{Sequential Learning Algorithm}\label{alg:seqlearning}
	\begin{algorithmic}[1]
		\Require base model $\mathcal{M}$, abstracted model $\mathcal{M'}$, set of relevant variables $R$, mapping $a$, temperature $T$, trade-off $\lambda$, learning rate $\eta$, $n_e$ number of learning epochs
		\Ensure locally optimal collection $\alpha$ of abstraction maps $\alpha_{X'}$
		
		\State Instantiate the parameter set $\mathbf{W} = \{\}$
		\State Retrieve the set $\mathcal{J}$
		\State \For{$j = (X',Y') \in \mathcal{J}$}{
			Setup $\mathbf{\tilde{W}} = \{\}$  \Comment{Setup the params}
			
			\If{$W_{X'} \notin \mathbf{W}$}{
				Instantiate matrix $W_{X'}$ and add it to $\mathbf{\tilde{W}}$
				
				\Else{Retrieve $W_{X'}$ from $\mathbf{W}$}
				}

			\If{$W_{Y'} \notin \mathbf{W}$}{
				Instantiate matrix $W_{Y'}$ and add it to $\mathbf{\tilde{W}}$
				
				\Else{Retrieve $W_{Y'}$ from $\mathbf{W}$}
				}
			
			Instantiate $\mathcal{L}_1 = D_{JSD}(\tmp{W_{X'}}\mu; \nu \tmp{W_{Y'}})$ \Comment{Setup the first term in the loss} 
			
			Instantiate $\mathcal{L}_2 = \sum_{i}\left(1-\max_{j}\tmp{W_{X'}}_{ij}\right) + \sum_{i}\left(1-\max_{j}\tmp{W_{Y'}}_{ij}\right)$ \Comment{Setup the second term in the loss} 
			
			Assemble the loss $\mathcal{L} = \lambda \mathcal{L}_1 +  \mathcal{L}_2$ \Comment{Setup the loss}
			
			\For{$n_e$}{
				Optimize by gradient descent $\mathbf{\tilde{W}} = \mathbf{\tilde{W}} - \eta \frac{\partial \mathcal{L}(\mathbf{\tilde{W}})}{\partial\mathbf{\tilde{W}}}$ \Comment{Learn}
			}
			
			Add solution $\mathbf{\tilde{W}}$ to $\mathbf{W}$
		}
		
	\end{algorithmic}
\end{algorithm}

The sequential learning approach follows a similar paradigm as the independent learning approach, but instead of learning a same matrix $W_{X'}$ multiple times, it uses the first learned value; thus, no aggregation is required at the end of learning.

\section{Neural network structures}\label{app:NN}

Here we offer a representation of the joint approach and the other approaches as neural networks. Since all the algorithms rely on gradient descent to solve optimization problems connected to the collection of sub-problems specified by $\mathcal{J}$, we can visualize them as neural networks.

In all our depictions, we will depict neural networks for solving an abstraction learning problem with the illustrative set $\mathcal{J} = \{ (X',Y'), (X',Z'), (Y',Z') \}$. We will use the following notation: circles represent known interventional distributions encoded in the form of stochastic matrices; solid rectangles represent neural network layers in which we instantiate the learnable parameters $W$ and we process inputs multiplying them by $\tmp{W}$; finally, the dashed box contains the loss function which produces the learning signal for gradient descent. We use color to highlight the paths belonging to each individual diagram: diagram of $(X',Y')$ in red, diagram of $(X',Z')$ in black, and diagram of $(Y',T')$ in blue.

\subsection{Independent approach}
\begin{figure}
\begin{center}
    
\begin{tikzpicture}[node distance={15mm}, thick, main/.style = {draw, circle}] 

\begin{scope}[xshift=-120]
\node[rectangle,dashed,thick,draw=black!75] (1) {$\lambda\mathcal{L}_{1} + \mathcal{L}_{2}$}; 
\node[main] (2) [below left of=1] {$\nu$};
\node[rectangle,thick,draw=black!75] (3) [below right of=1] {$W_{Y'}$};
\node[main] (4) [below of=3] {$\mu$};
\node[rectangle,thick,draw=black!75] (5) [below of=2] {$W_{X'}$};
\node[] (6) [below right of=5] {$do(X)$};

\draw[->,draw=red] (6) -- (4);
\draw[->,draw=red] (6) -- (5);
\draw[->,draw=red] (5) -- (2);
\draw[->,draw=red] (4) -- (3);
\draw[->,draw=red] (3) -- (1);
\draw[->,draw=red] (2) -- (1);
\end{scope}

\begin{scope}[xshift=0]
\node[rectangle,dashed,thick,draw=black!75] (1) {$\lambda\mathcal{L}_{1} + \mathcal{L}_{2}$}; 
\node[main] (2) [below left of=1] {$\nu'$};
\node[rectangle,thick,draw=black!75] (3) [below right of=1] {$W_{Z'}$};
\node[main] (4) [below of=3] {$\mu'$};
\node[rectangle,thick,draw=black!75] (5) [below of=2] {$W_{Y'}$};
\node[] (6) [below right of=5] {$do(Y)$};

\draw[->,draw=blue] (6) -- (4);
\draw[->,draw=blue] (6) -- (5);
\draw[->,draw=blue] (5) -- (2);
\draw[->,draw=blue] (4) -- (3);
\draw[->,draw=blue] (3) -- (1);
\draw[->,draw=blue] (2) -- (1);
\end{scope}

\begin{scope}[xshift=120]
\node[rectangle,dashed,thick,draw=black!75] (1) {$\lambda\mathcal{L}_{1} + \mathcal{L}_{2}$}; 
\node[main,text width=1cm,minimum size=1cm,align=center,inner sep=0pt] (2) [below left of=1] {$\nu' \circ \nu$};
\node[rectangle,thick,draw=black!75] (3) [below right of=1] {$W_{Z'}$};
\node[main,text width=1cm,minimum size=1cm,align=center,inner sep=0pt] (4) [below of=3] {$\mu' \circ \mu$};
\node[rectangle,thick,draw=black!75] (5) [below of=2] {$W_{X'}$};
\node[] (6) [below right of=5] {$do(X)$};
\draw[->,draw=black] (6) -- (4);
\draw[->,draw=black] (6) -- (5);
\draw[->,draw=black] (5) -- (2);
\draw[->,draw=black] (4) -- (3);
\draw[->,draw=black] (3) -- (1);
\draw[->,draw=black] (2) -- (1);
\end{scope}

\end{tikzpicture}
\end{center}
\caption{Neural network structure implied by the independent approach. Three separated and independent neural networks are instantiated, each one learning its own parameters. At the end two copies of $W_{X'},W_{Y'},W_{Z'}$ will be learned.}
\label{fig:app_nn_indep}
\end{figure}
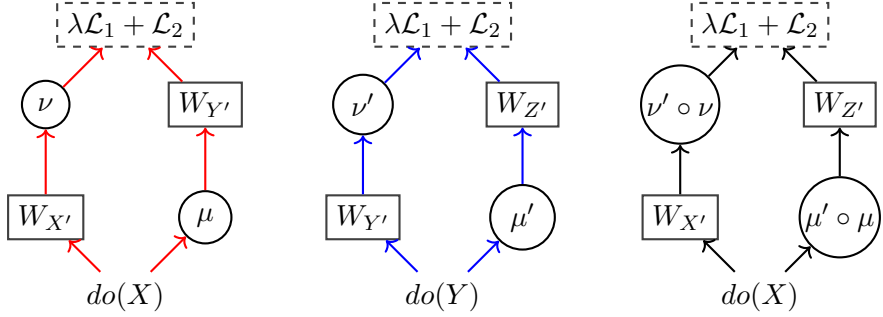

Figure \ref{fig:app_nn_indep} shows the structure defined by the independent approach. This approach solves the sub-problems defined by $\mathcal{J}$ independently; therefore, this may be seen as instantiating an independent neural network for each diagram.

\subsection{Sequential approach}
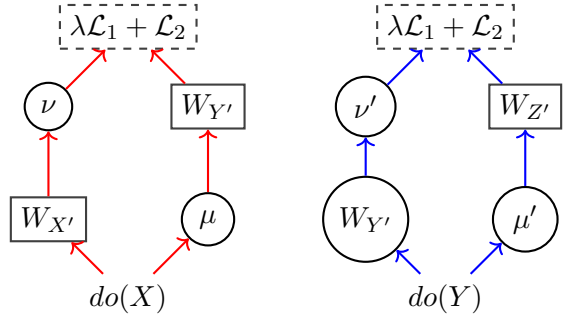
\begin{figure}
\begin{center}
    
\begin{tikzpicture}[node distance={15mm}, thick, main/.style = {draw, circle}] 

\begin{scope}[xshift=-120]
\node[rectangle,dashed,thick,draw=black!75] (1) {$\lambda\mathcal{L}_{1} + \mathcal{L}_{2}$}; 
\node[main] (2) [below left of=1] {$\nu$};
\node[rectangle,thick,draw=black!75] (3) [below right of=1] {$W_{Y'}$};
\node[main] (4) [below of=3] {$\mu$};
\node[rectangle,thick,draw=black!75] (5) [below of=2] {$W_{X'}$};
\node[] (6) [below right of=5] {$do(X)$};

\draw[->,draw=red] (6) -- (4);
\draw[->,draw=red] (6) -- (5);
\draw[->,draw=red] (5) -- (2);
\draw[->,draw=red] (4) -- (3);
\draw[->,draw=red] (3) -- (1);
\draw[->,draw=red] (2) -- (1);
\end{scope}

\begin{scope}[xshift=0]
\node[rectangle,dashed,thick,draw=black!75] (1) {$\lambda\mathcal{L}_{1} + \mathcal{L}_{2}$}; 
\node[main] (2) [below left of=1] {$\nu'$};
\node[rectangle,thick,draw=black!75] (3) [below right of=1] {$W_{Z'}$};
\node[main] (4) [below of=3] {$\mu'$};
\node[main] (5) [below of=2] {$W_{Y'}$};
\node[] (6) [below right of=5] {$do(Y)$};

\draw[->,draw=blue] (6) -- (4);
\draw[->,draw=blue] (6) -- (5);
\draw[->,draw=blue] (5) -- (2);
\draw[->,draw=blue] (4) -- (3);
\draw[->,draw=blue] (3) -- (1);
\draw[->,draw=blue] (2) -- (1);
\end{scope}

\end{tikzpicture}
\end{center}
\caption{Neural network structure implied by the sequential approach. Only two neural networks are instantiated in order to learn $W_{X'},W_{Y'},W_{Z'}$. Notice how, in the second network, the value $W_{Y'}$ is not a learnable parameter anymore (square box) but a fixed matrix (circle).} \label{fig:app_nn_seq}
\end{figure}

Figure \ref{fig:app_nn_seq} shows the structure defined by the sequential approach. This approach solves the sub-problems defined by $\mathcal{J}$ independently, but it avoids re-learning weight matrices; this may be seen as instantiating the minimal number of independent neural network to learn all the abstraction maps.

\subsection{Joint approach}
Figure \ref{fig:nn} shows the structure defined by the joint approach. Notice how the individual neural networks defined by the independent or sequential approach get merged into a single neural network with its internal connections and backpropagation defined by $\mathcal{J}$.

\section{Details about the synthetic-data simulations} \label{app:SyntheticData}

\subsection{Synthetic models}\label{app:SyntheticData_models}

Here we provide details about our synthetic models. For each abstraction learning scenario, we define the two SCMs $\mathcal{M}, \mathcal{M'}$ by listing the set of nodes and their domains; we define the abstraction by listing the set $R$ of relevant nodes and the structural-level map $a$. We illustrate mechanisms and abstractions in a figure, where we follow the convention of representing the base model on top, and the abstracted model on the bottom. We express mechanisms and abstractions as matrices in the figure. A matrix over a solid edge with no source and target $X$ represents distribution $P(X)$. A matrix next to one or more solid edges with sources $X_1,X_2,...,X_m$ and target $Y$ represents conditional distribution $P(Y \vert X_1,X_2,...,X_m)$. A matrix along a dashed edge with target $X'$ represents abstraction $\alpha_{X'}$.

\subsubsection{Basic lung cancer scenario}

This example is taken from \cite{rischel2020category}, and represent a toy lung cancer scenario defined over the variables \emph{smoking} ($S,S'$), \emph{tar deposits in the lungs} ($T$), and \emph{lung cancer} ($C,C'$).
The base and the abstracted models are defined on the following nodes:
\begin{center}
\begin{tabular}{ccc}
	$\mathcal{X}=\left\{ S,T,C\right\} $ & \;\; & $\mathcal{X}'=\left\{ S',C'\right\} $\tabularnewline
	$\mathcal{M}[S]=\mathcal{M}[T]=\mathcal{M}[C]=\left\{ 0,1\right\} $ &  & $\mathcal{M}'[S']=\mathcal{M}'[C']=\left\{ 0,1\right\} $\tabularnewline
\end{tabular}
\end{center}

Abstraction is (partially) specified as:
\begin{center}
\begin{tabular}{ccc}
	& $R=\left\{ S,C\right\} $ & \tabularnewline
	& $a(S)=S'$, $a(C)=C'$ & \tabularnewline
\end{tabular}
\end{center}

The following figure reports mechanisms and optimal abstractions:

\begin{center}
\begin{tikzpicture}[shorten >=1pt, auto, node distance=1cm, thick, scale=0.8, every node/.style={scale=0.8}]
	\tikzstyle{node_style} = [circle,draw=black]
	\node[node_style] (S) at (0,0) {S};
	\node[node_style] (T) at (2.5,0) {T};
	\node[node_style] (C) at (5,0) {C};
	
	\node[node_style] (Si) at (0,-3) {S'};
	\node[node_style] (Ci) at (5,-3) {C'};
	\draw[->]  (-2,0) to node[above,font=\small]{$\left[\begin{array}{c}
			.8\\
			.2
		\end{array}\right]$} (S);
	\draw[->]  (S) to node[above,font=\small]{$\left[\begin{array}{cc}
			1 & .2\\
			0 & .8
		\end{array}\right] $} (T);
	\draw[->]  (T) to node[above,font=\small]{$\left[\begin{array}{cc}
			.9 & .6 \\
			.1 & .4 
		\end{array}\right]$} (C);
	
	\draw[->]  (-2,-3) to node[below,font=\small]{$\left[\begin{array}{c}
			.8\\
			.2
		\end{array}\right]$} (Si);
	\draw[->]  (Si) to node[below,font=\small]{$\left[\begin{array}{cc}
			.9 & .66\\
			.1 & .34
		\end{array}\right]$} (Ci);
	
	\draw[->,dashed]  (S) to node[left,font=\small]{$\left[\begin{array}{cc}
			1 & 0\\
			0 & 1
		\end{array}\right]$} (Si);
	\draw[->,dashed]  (C) to node[right,font=\small]{$\left[\begin{array}{cc}
			1 & 0\\
			0 & 1
		\end{array}\right]$} (Ci);
\end{tikzpicture}
\end{center}

We consider as relevant interventional distributions:
\begin{center}
\begin{tabular}{ccc}
	& $P'(C'\vert do(S')),$ &
\end{tabular}
\end{center}
and, therefore, $\mathcal{J} = \{ (S',C') \}$.

With respect to this $\mathcal{J}$, the abstraction has an optimal zero-error $e(\abs)=0$ computed by enumeration. Enumeration is feasible since, according to Equation \ref{eqn:all_n_surj_functions}, the number of solutions to be evaluated are:
\begin{equation}
    |\mathcal{A}| = 2! \iiks{2}{2} \cdot 2! \iiks{2}{2} = 4.
\end{equation}

\subsubsection{Collapsing lung cancer scenario}

This scenario enriches the base model with a new variable accounting for \emph{environmental factors} ($E$), and considers an abstraction in which this variable is marginalized away.
The base and the abstracted models are defined on the following nodes:

\begin{center}
	\begin{tabular}{ccc}
		$\mathcal{X}=\left\{ E,S,T,C\right\} $ & \;\;  & $\mathcal{X}'=\left\{ S',T',C'\right\} $\tabularnewline
		$\mathcal{M}[E]=\mathcal{M}[S]=\mathcal{M}[T]=\mathcal{M}[C]=\left\{ 0,1\right\} $ &  & $\mathcal{M}'[S']=\mathcal{M}'[T']=\mathcal{M}'[C']=\left\{ 0,1\right\} $\tabularnewline
	\end{tabular}
\end{center}

Abstraction is (partially) specified as:
\begin{center}	
	\begin{tabular}{ccc}
		& $R=\left\{ E,S,T,C\right\} $ & \tabularnewline
		& $a(E)=S', a(S)=S', a(T)=T', a(C)=C'$ & \tabularnewline
	\end{tabular}
\end{center}

The following figure reports mechanisms and optimal abstractions:

\begin{center}
	\begin{tikzpicture}[shorten >=1pt, auto, node distance=1cm, thick, scale=0.8, every node/.style={scale=0.8}]
		\tikzstyle{node_style} = [circle,draw=black]
		\node[node_style] (E) at (-2.5,0) {E};
		\node[node_style] (S) at (0,0) {S};
		\node[node_style] (T) at (2.5,0) {T};
		\node[node_style] (C) at (5,0) {C};
		
		\node[node_style] (Si) at (-1,-3) {S'};
		\node[node_style] (Ti) at (2.5,-3) {T'};
		\node[node_style] (Ci) at (5,-3) {C'};
		\draw[->]  (-4.5,0) to node[above,font=\small]{$\left[\begin{array}{c}
				.45\\
				.55
			\end{array}\right]$} (E);
		\draw[->]  (E) to node[above,font=\small]{$\left[\begin{array}{cc}
				.9 & .7\\
				.1 & .3
			\end{array}\right] $} (S);
		\draw[->]  (S) to node[above,font=\small]{$\left[\begin{array}{cc}
				.95 & .2\\
				.05 & .8
			\end{array}\right] $} (T);
		\draw[->]  (T) to node[above,font=\small]{$\left[\begin{array}{cc}
				.9 & .6 \\
				.1 & .4 
			\end{array}\right]$} (C);
		
		\draw[->]  (-3,-3) to node[below,font=\small]{$\left[\begin{array}{c}
				.8\\
				.2
			\end{array}\right]$} (Si);
		\draw[->]  (Si) to node[below,font=\small]{$\left[\begin{array}{cc}
				1 & .2\\
				0 & .8
			\end{array}\right]$} (Ti);
		\draw[->]  (Ti) to node[below,font=\small]{$\left[\begin{array}{cc}
				.9 & .6\\
				.1 & .4
			\end{array}\right]$} (Ci);
		
		\draw[->,dashed]  (E) to node[left,font=\small]{$\left[\begin{array}{cccc}
				1 & 0 & 1 & 0\\
				0 & 1 & 0 & 1
			\end{array}\right]$} (Si);
		\draw[->,dashed]  (S) to (Si);
		\draw[->,dashed]  (T) to node[left,font=\small]{$\left[\begin{array}{cc}
				1 & 0\\
				0 & 1
			\end{array}\right]$} (Ti);
		\draw[->,dashed]  (C) to node[right,font=\small]{$\left[\begin{array}{cc}
				1 & 0\\
				0 & 1
			\end{array}\right]$} (Ci);
	\end{tikzpicture}
\end{center}

We consider as relevant interventional distributions:
\begin{center}
\begin{tabular}{ccc}
	& $P'(C'\vert do(S')),$ & \tabularnewline
	& $P'(C'\vert do(T')),$ & \tabularnewline
	& $P'(T'\vert do(S')),$ & \tabularnewline
\end{tabular}
\end{center}
and, therefore, $\mathcal{J} = \{ (S',C'), (T',C'), (S',T') \}$.

With respect to this $\mathcal{J}$, the abstraction has an optimal non-zero-error $e(\abs)\approx0.13$ computed by enumeration. Enumeration is feasible since, according to Equation \ref{eqn:all_n_surj_functions}, the number of solutions to be evaluated are:
\begin{equation}
    |\mathcal{A}| = 2! \iiks{4}{2} \cdot 2! \iiks{2}{2} \cdot 2! \iiks{2}{2} = 56.
\end{equation}

\subsubsection{Extended lung cancer scenario}

This example constructs a scenario where variables have different cardinalities, and abstraction is focused on reducing the resolution of the experiments.
The base and the abstracted models are defined on the following nodes:
\begin{center}
	\begin{tabular}{ccc}
		$\mathcal{X}=\left\{ S,T,C\right\} $ &  \;\; & $\mathcal{X}'=\left\{ S',T',C'\right\} $\tabularnewline
		$\mathcal{M}[S]=\left\{ 0,1,2,3\right\} $ &  & $\mathcal{M}'[S']=\left\{ 0,1,2\right\} $\tabularnewline
		$\mathcal{M}[T]=\left\{ 0,1,2\right\} $ &  & $\mathcal{M}'[T']=\left\{ 0,1\right\} $\tabularnewline
		$\mathcal{M}[C]=\left\{ 0,1\right\} $ &  & $\mathcal{M}'[C']=\left\{ 0,1\right\} $\tabularnewline
	\end{tabular}
\end{center}

Abstraction is (partially) specified as:
\begin{center}	
	\begin{tabular}{ccc}
		& $R=\left\{ S,T,C\right\} $ & \tabularnewline
		& $a(S)=S', a(T)=T',a(C)=C'$ & \tabularnewline
	\end{tabular}
\end{center}

The following figure reports mechanisms and optimal abstractions:

\begin{center}
	\begin{tikzpicture}[shorten >=1pt, auto, node distance=1cm, thick, scale=0.8, every node/.style={scale=0.8}]
		\tikzstyle{node_style} = [circle,draw=black]
		\node[node_style] (S) at (0,0) {S};
		\node[node_style] (T) at (3.5,0) {T};
		\node[node_style] (C) at (7,0) {C};
		
		\node[node_style] (Si) at (0,-3) {S'};
		\node[node_style] (Ti) at (3.5,-3) {T'};
		\node[node_style] (Ci) at (7,-3) {C'};
		\draw[->]  (-2,0) to node[above,font=\small]{$\left[\begin{array}{c}
				.25\\
				.25 \\
				.25 \\
				.25
			\end{array}\right]$} (S);
		\draw[->]  (S) to node[above,font=\small]{$\left[\begin{array}{cccc}
				.6 & .55 & .1 & .1 \\
				.3 & .25 & .4 & .4 \\
				.1 & .2 & .5 & .5
			\end{array}\right] $} (T);
		\draw[->]  (T) to node[above,font=\small]{$\left[\begin{array}{ccc}
				.7 & .7 & .4 \\
				.3 & .3 & .6
			\end{array}\right]$} (C);
		
		\draw[->]  (-2,-3) to node[below,font=\small]{$\left[\begin{array}{c}
				.25\\
				.5 \\
				.25
			\end{array}\right]$} (Si);
		\draw[->]  (Si) to node[below,font=\small]{$\left[\begin{array}{ccc}
				.9 & .8 & .5\\
				.1 & .2 & .5
			\end{array}\right]$} (Ti);
		\draw[->]  (Ti) to node[below,font=\small]{$\left[\begin{array}{cc}
				.7 & .4\\
				.3 & .6
			\end{array}\right]$} (Ci);
		
		\draw[->,dashed]  (S) to node[left,font=\small]{$\left[\begin{array}{cccc}
				1 & 0 & 0 & 0\\
				0 & 1 & 0 & 0\\
				0 & 0 & 1 & 1
			\end{array}\right]$} (Si);
		\draw[->,dashed]  (T) to node[left,font=\small]{$\left[\begin{array}{ccc}
				1 & 1 & 0\\
				0 & 0 & 1
			\end{array}\right]$} (Ti);
		\draw[->,dashed]  (C) to node[right,font=\small]{$\left[\begin{array}{cc}
				1 & 0\\
				0 & 1
			\end{array}\right]$} (Ci);
	\end{tikzpicture}
\end{center}	

We consider as relevant interventional distributions:
\begin{center}
\begin{tabular}{ccc}
	& $P'(C'\vert do(S')),$ & \tabularnewline
	& $P'(C'\vert do(T')),$ & \tabularnewline
	& $P'(T'\vert do(S')),$ & \tabularnewline
\end{tabular}
\end{center}
and, therefore, $\mathcal{J} = \{ (S',C'), (T',C'), (S',T') \}$.

With respect to this $\mathcal{J}$, the abstraction has an optimal zero-error $e(\abs)=0$ computed by enumeration. Enumeration is feasible since, according to Equation \ref{eqn:all_n_surj_functions}, the number of solutions to be evaluated are:
\begin{equation}
    |\mathcal{A}| = 3! \iiks{4}{3} \cdot 2! \iiks{3}{2} \cdot 2! \iiks{2}{2} = 432.
\end{equation}

\subsubsection{v-Structure lung cancer scenario}

This scenario introduces more complex v-structures, and is partly inspired by the LUCAS toydaset\footnote{\url{http://www.causality.inf.ethz.ch/data/LUCAS.html}}. Beyond the previous variables, new observables are introduced, such as \emph{genetic factors} ($G$), \emph{coughing} ($H$), and \emph{fatigue} ($F$).
The base and the abstracted models are defined on the following nodes:
\begin{center}
	\begin{tabular}{ccc}
		$\mathcal{X}=\left\{ S,G,C,H,F\right\} $ & \;\;  & $\mathcal{X}'=\left\{ S',C',F'\right\} $\tabularnewline
		$\mathcal{M}[S]=\mathcal{M}[G]=\mathcal{M}[C]=\mathcal{M}[J]=\mathcal{M}[F]=\left\{ 0,1\right\} $ &  & $\mathcal{M}'[S']=\mathcal{M}'[C']=\mathcal{M'}[F']=\left\{ 0,1\right\} $\tabularnewline
	\end{tabular}
\end{center}

Abstraction is (partially) specified as:
\begin{center}	
	\begin{tabular}{ccc}
		& $R=\left\{ S,C,H,F\right\} $ & \tabularnewline
		& $a(S)=S'$, $a(C)=C', a(H)=F', a(F)=F'$ & \tabularnewline
	\end{tabular}
\end{center}

The following figure reports mechanisms and optimal abstractions:

\begin{center}
	\begin{tikzpicture}[shorten >=1pt, auto, node distance=1cm, thick, scale=0.8, every node/.style={scale=0.8}]
		\tikzstyle{node_style} = [circle,draw=black]
		\node[node_style] (S) at (0,-1) {S};
		\node[node_style] (G) at (0,1) {G};
		\node[node_style] (C) at (4.5,0) {C};
		\node[node_style] (H) at (9,1) {H};
		\node[node_style] (F) at (9,-1) {F};
		
		\node[node_style] (Si) at (0,-4) {S'};
		\node[node_style] (Ci) at (4.5,-4) {C'};
		\node[node_style] (Fi) at (9,-4) {F'};
		
		\draw[->]  (-2,-1) to node[above,font=\small]{$\left[\begin{array}{c}
				.8\\
				.2
			\end{array}\right]$} (S);
		\draw[->]  (-2,1) to node[above,font=\small]{$\left[\begin{array}{c}
				.7\\
				.3
			\end{array}\right]$} (G);
		
		\draw[->]  (S) to (C);
		\draw[->]  (G) to node[above,font=\small]{$\left[\begin{array}{cccc}
				.15 & .85 & .65 & .75\\
				.85 & .15 & .35 & .25
			\end{array}\right]$} (C);
		\draw[->]  (C) to node[above,font=\small]{$\left[\begin{array}{cccc}
				1 & .2 \\
				0 & .8 
		\end{array}\right]$} (H);
		\draw[->]  (C) to (F);
		\draw[->]  (H) to node[right,font=\small]{$\left[\begin{array}{cccc}
				.42 & .75 & .65 & .33\\
				.58 & .25 & .35 & .67
			\end{array}\right]$} (F);

		\draw[->]  (-2,-4) to node[below,font=\small]{$\left[\begin{array}{c}
				.8\\
				.2
			\end{array}\right]$} (Si);
		\draw[->]  (Si) to node[below,font=\small]{$\left[\begin{array}{cc}
				.9 & .66\\
				.1 & .34
			\end{array}\right]$} (Ci);
		\draw[->]  (Ci) to node[below,font=\small]{$\left[\begin{array}{cc}
				.8 & .5\\
				.2 & .5
			\end{array}\right]$} (Fi);
		
		\draw[->,dashed]  (S) to node[left,font=\small]{$\left[\begin{array}{cc}
				0 & 1\\
				1 & 0
			\end{array}\right]$} (Si);
		\draw[->,dashed]  (C) to node[right,font=\small]{$\left[\begin{array}{cc}
				1 & 0\\
				0 & 1
			\end{array}\right]$} (Ci);
		\draw[->,dashed,bend right]  (H) to (Fi);
		\draw[->,dashed]  (F) to node[right,font=\small]{$\left[\begin{array}{cccc}
				0 & 1 & 0 & 1\\
				1 & 0 & 1 & 0
			\end{array}\right]$} (Fi);
	\end{tikzpicture}
\end{center}

We consider as relevant interventional distributions:
\begin{center}
\begin{tabular}{ccc}
	& $P'(C'\vert do(S')),$ & \tabularnewline
	& $P'(F'\vert do(C')),$ & \tabularnewline
	& $P'(F'\vert do(S')),$ & \tabularnewline
\end{tabular}
\end{center}
and, therefore, $\mathcal{J} = \{ (S',C'), (C',F'), (S',F') \}$.

With respect to this $\mathcal{J}$, the abstraction has an optimal non-zero-error $e(\abs) \approx 0.21$ computed by enumeration. Enumeration is feasible since, according to Equation \ref{eqn:all_n_surj_functions}, the number of solutions to be evaluated are:
\begin{equation}
    |\mathcal{A}| = 2! \iiks{2}{2} \cdot 2! \iiks{2}{2} \cdot 2! \iiks{4}{2} = 56.
\end{equation}

\subsection{Experimental settings} \label{app:SyntheticData_experimental}

Here we provide details about the experimental settings for all the simulations on synthetic data.

\subsubsection{Comparison with baselines}

In simulation (a) we learn an abstraction on each one of the four synthetic scenarios (\emph{basic, collapsing, extended, v-structure}) using three approaches:
\begin{itemize}
    \item \emph{Independent approach}: we use the algorithm presented in Appendix \ref{app:IndependentApproach}. As an aggregation technique (line 4 in Algorithm \ref{alg:indeplearning}) we use the following algorithm: given multiple solutions $W'_{X'}, W''_{X'}, W'''_{X'}, ..., W^{m}_{X'}$ for $W_{X'}$, for each column $i$ in $W_{X'}$, we select the index with the highest value among all the candidates, that is, $j^* = argmax_j \left\{ {W'_{X'}}_{ji}, {W''_{X'}}_{ji}, {W'''_{X'}}_{ji} ... , {W^{m}_{X'}}_{ji}\right\}$; we then generate a new aggregated solution $\bar{W}_{X'}$ where column $i$ has a one in position $j^*$, while all the other values in the column are zero.
    \item \emph{Sequential approach}: we use the algorithm presented in Appendix \ref{app:SequentialApproach}; we adopt a random ordering of the diagrams in $\mathcal{J}$. 
    \item \emph{Joint approach}: we use the algorithm presented in Appendix \ref{app:JointApproach}.
\end{itemize}

All algorithms are run with the same settings: temperature $T=0.1$, trade-off $\lambda=10$, learning rate $\eta=0.01$, number of epochs $n_e = 500$. We optimize by gradient descent using the Adam algorithm \citep{kingma2014adam}. For each approach, we run an ensemble of 10 models, and we select the best wrt the JSD loss term $\mathcal{L}_1$. We repeat experiments 10 times in order to collect reliable statistics.

We measure the performance of the three approaches using different metrics. First of all, we define a set of normalized metrics used to monitor learning at runtime:
\begin{itemize}
    \item \emph{Normalized JSD loss}: $\frac{1}{|\mathcal{J}_{train}|} \sum_{(\mathbf{X'},\mathbf{Y'}) \in \mathcal{J}_{train}} D_{JSD}(\alpha_{\mathbf{Y'}}  \mu; \nu  \alpha_{\mathbf{X'}})$; this corresponds to the sum of JSDs on each diagram considered during training, divided by the number of such diagrams. Normalization is required because, during training, the independent and joint approach consider all the diagrams $\mathcal{J}_{train} = \mathcal{J}$, but the sequential approach may consider a smaller  number of diagrams $\mathcal{J}_{train} \subseteq \mathcal{J}$ (i.e.: the sequential approach may ignore a diagram defined over variables $\alpha_{X'}$ already computed in other diagrams).
    
    \item \emph{Normalized surjective penalty}: $\frac{1}{|\mathbf{W}_{train}|}\sum_{W\in\mathbf{W}_{train}}\sum_{i=1}^{M_{i}}\left(1-\max_{j}\tmp{W}_{ij} \right)$; this corresponds to the sum of surjective penalties on each weight matrix instantiated during training, divided by the number of such matrices. Normalization is required because, during training, the surjective and joint approach instantiate a number of weight matrices equal to $|\mathbf{W}_{train}|=|\mathcal{X'}|$, but the independent approach may instantiate the same weight matrix multiple times $|\mathbf{W}_{train}|\geq|\mathcal{X'}|$ (i.e.: the independent approach may solve different diagrams in the same $\alpha_{X'}$ independently at the same time).
    
    \item \emph{Normalized L1 distance}: $\frac{1}{|\mathbf{W}_{train}|}\sum_{W\in\mathbf{W}_{train}} \lone(W - W^*)$, where $\lone(W) = \sum_{i,j} |W_{ij}|$ is the $\lone$-norm, and $W^*$ is the ground-truth optimal solution computed via enumeration; this corresponds to the sum of the $\lone$ distances from the optimal solution of the current weight matrices, divided by the number of such matrices. As in the case of surjective penalty, normalization is required because, during training, a different number of weight matrices may be instantiated by each approach.
\end{itemize} 

At the end of the training, all the approaches return the collection of learned maps $\mathbf{W}$. We discretize the solutions in $\{0,1\}$ by rounding, and we evaluate the quality of the computed solution using the following metrics:

\begin{itemize}
    \item \emph{JSD loss}: $ \sum_{(\mathbf{X'},\mathbf{Y'}) \in \mathcal{J}} D_{JSD}(\alpha_{\mathbf{Y'}}  \mu; \nu  \alpha_{\mathbf{X'}})$; this corresponds to the sum of JSDs on all the diagrams in $|\mathcal{J}|$.

    \item \emph{Surjective penalty}: $\sum_{W\in\mathbf{W}}\sum_{i=1}^{M_{i}}\left(1-\max_{j}\tmp{W}_{ij} \right)$; this corresponds to the sum of surjective penalties on all weight matrices in $\mathbf{W}$. 
    
    \item \emph{L1 distance}: $\sum_{W\in\mathbf{W}} \lone (W - W^*)$; this corresponds to the sum of the $\lone$ distances from the optimal solution of all the weight matrices $\mathbf{W}$.
    
    \item \emph{Wallclock time}: we provide an estimate of the running time of each approach. Since our implementation is not parallel, we divide the runtime of the independent approach by the number of diagrams, to simulate the possibility of running each sub-problem in parallel.
\end{itemize}

\subsubsection{Evaluation of weighting}

In simulation (b) we learn an abstraction in the  \emph{v-structure} scenario. In this scenario, we have defined the set of relevant interventional distribution as $\mathcal{J} = \{ (S',C'), (C',F'), (S',F') \}$. We now assume that one intervention (how intervening on smoking affects cancer) is more important than the others (how intervening on smoking affects fatigue, or how intervening on cancer affects fatigue); we also assume that, for the same reason, we have collected a larger number of samples for $P'(C'\vert do(S'))$. We then impose a weighting schema $\boldsymbol{\kappa}= [2.4,0.3,0.3]$ that scales the loss wrt intervention $P'(C'\vert do(S'))$ more heavily than wrt the other interventions. Notice that the weighting values are chosen to sum up to the same value that the standard uniform weighting $[1,1,1]$ would have; keeping the same magnitude is important not to change the ratio between the JSD loss and the surjective penalty.

We run only our algorithm (\emph{joint approach}), with the same setting as in simulation (a).

We measure the quality of the result at the end of training by focusing on \emph{JSD loss}, and comparing the solution to the result obtained without weighting.

\subsubsection{Ablation study}

In simulation (c) we learn an abstraction in the  \emph{v-structure} scenario without using the surjectivity penalty term $\mathcal{L}_2$ in the loss function.

We run only our algorithm (\emph{joint approach}), with the same setting as in simulation (a).

We measure the quality of the result at the end of training by focusing on \emph{JSD loss} and \emph{L1 distance}, and comparing the solution to the result obtained with the penalty term.

\subsubsection{Sensitivity analysis}

In simulation (d) we analyze how abstraction learning changes as a function of the hyperparameters $T$ and $\lambda$ specific to our algorithm.

We run only our algorithm (\emph{joint approach}), with the same setting as in simulation (a), except for the hyperparameter $T$ which is chosen in the set $\{0.01, 0.05, 0.1, 0.5, 1\}$ and the parameter $\lambda$ which is chosen in the set $\{1, 5, 10, 20, 50\}$.

We measure the quality of the result at runtime by considering \emph{L1 distance} at the end of training.

\subsection{Additional results}\label{app:SyntheticData_additional}

\begin{figure}
\resizebox{\columnwidth}{!}{%
\includegraphics{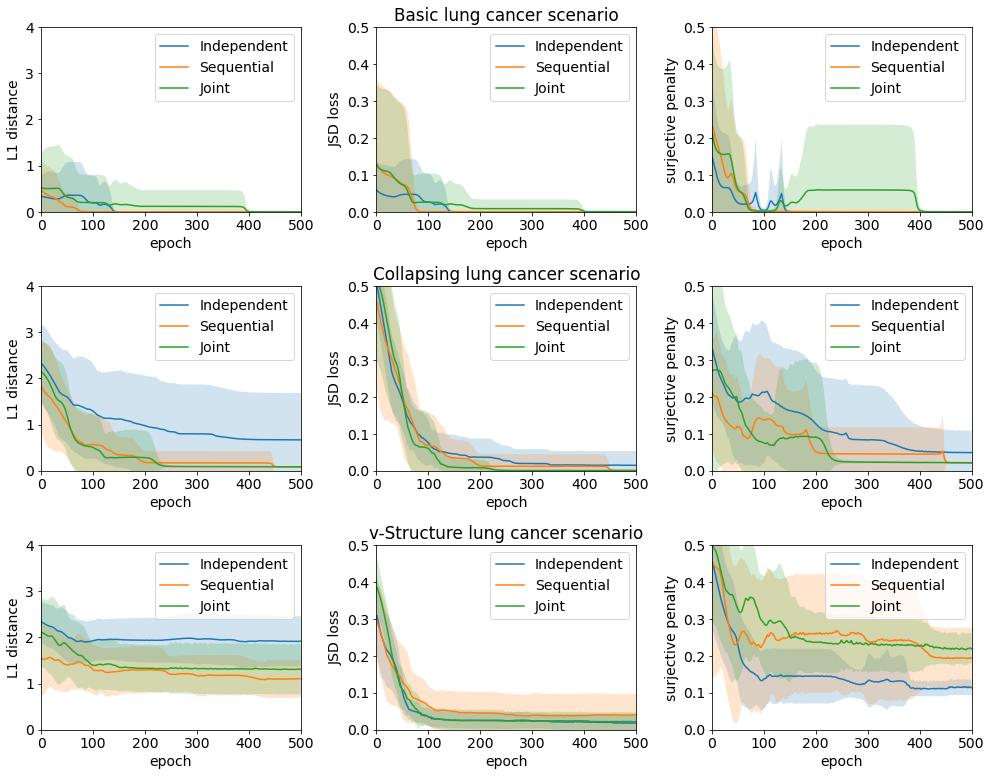}}
\caption{Performance during training on the \emph{basic}, \emph{collapsing}, and \emph{v-structure} scenarios.}
\label{fig:syntexp_a_performances}
\end{figure}

Figure \ref{fig:syntexp_a_performances} provides results of simulation (a) on all the remaining scenarios. In terms of $\lone$-distance from the ground truth, the joint approach achieves performances in line or better than the other approaches.

\begin{figure}
\begin{minipage}[t]{0.65\linewidth}\centering
\resizebox{\columnwidth}{!}{%
\includegraphics{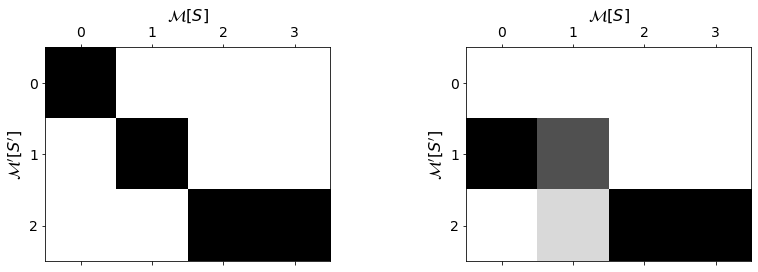}}
\caption{Learned matrix $W_{S'}$ with surjective penalty (left) and without (right).}
\label{fig:syntexp_c_learned_Ws}
\end{minipage}\hfill%
\begin{minipage}[t]{0.32\linewidth}\centering
\resizebox{\columnwidth}{!}{%
\includegraphics{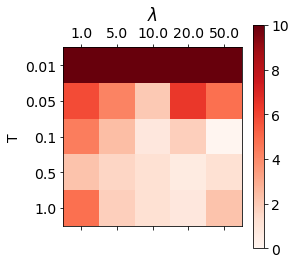}}
\caption{Performance as $\lone$-distance when using different configuration of $T$ and $\lambda$.}
\label{fig:syntexp_d_performance}
\end{minipage}
\end{figure}

Figure \ref{fig:syntexp_c_learned_Ws} shows sample matrices $W_{S'}$ learned with and without surjective penalty. Without surjectivity penalty, $W_{S'}$ simply ignore the value $S'=0$ in $\mathcal{M'}$ since no value from the base model would be mapped onto it. This could allow the algorithm to achieve a better JSD by reducing the support on which the error is computed. In the limit case, the algorithm may map all the values in the base model onto a single value in $\mathcal{M'}$ in order to reduce JSD. Such a solution would be meaningless, and a surjective penalty prevent this form of collapse.

Figure \ref{fig:syntexp_d_performance} shows the result of our sensitivity study. In general, the joint algorithm seems to produce reliable results for different combinations of the temperature parameter $T$ and the trade-off parameter $\lambda$. We notice, however, that learning is hindered for values of $T\leq 0.05$; this is likely due to an excessively low temperature that prevents a continuous relaxation of the solution space sufficiently smooth to be explored by gradient descent. Low value of trade-off may also have a negative impact on learning; this should likely be ascribed by the surjectivity penalty becoming the leading factor in learning, overshadowing the contribution of JSD.

\section{Details about the real-world-data simulations} \label{app:RealData}

\subsection{Data and models} \label{app:RealData_models}

Here we provide details about the real-world data we used, and the models we devised. We use the same graphical conventions used in the Appendix \ref{app:SyntheticData_models}.

\subsubsection{LRCS dataset}
The LRCS dataset records the results of a set of experiments investigating the relationship between lithium-ion battery manufacturing parameters for slurry and coating, and target parameters \citep{cunha2020artificial}. This dataset is publicly available at \url{https://chemistry-europe.onlinelibrary.wiley.com/doi/full/10.1002/batt.201900135}.

Each sample in the LRCS dataset represents the result of an experiment, and it is defined by four features:
\begin{itemize}
    \item \emph{AM Composition:} composition of the slurry as a percentage of active material. This parameter assumes values in $\{92.7, 94 , 95 , 96\}$;
    
    \item \emph{S-to-L ratio:} solid-to-liquid ratio of the slurry as a percentage. This parameter assumes continuous values in $(54,75)$;
    
    \item \emph{Comma gap:} gap in the coating process, measured in $[\mu m]$. This parameter assumes discrete values in $\{50, 75 , 100, 200, 300, 400\}$;
    
    \item \emph{Viscosity:} viscosity of the slurry, measured in $[Pa \cdot s]$. This parameter assumes continuous values in $(1,14)$.
\end{itemize}

Moreover, for each experiment, two target variables are collected:
\begin{itemize}
    \item \emph{Mass loading:} mass loading of the coating, measured in $\left[ \frac{mg}{cm^2}\right]$. This variable assumes continuous values in $(4,54)$;
    
    \item \emph{Porosity:} porosity of the coating as a percentage. This parameter assumes continuous values in $(41,61)$.
\end{itemize}

The dataset contains 656 datapoints, that is, 8 datapoints for each one of 82 configurations of control parameters considered. 
%Following the approach in the literature \cite{liu2022interpretable}, we average the target variables in each configuration, thus reducing the dataset to 82 samples.

\subsubsection{WMG dataset}

The WMG group has collected a set of recordings from an experiment aimed at measuring the relationship between coating parameters and mass loading. This dataset is available at \url{https://github.com/mattdravucz/jointly-learning-causal-abstraction/}.

Each recording contains a large set of parameters defining the state of a coater machine. These parameters are sampled every second in the course of an experiment lasting about three hours. Three values are relevant to us:
\begin{itemize}
    \item \emph{Comma Bar Operator Position Actual:} basic gap in the coating process, measured in $[mm]$. This parameter, converted to  $[\mu m]$, assumes discrete values  in $[0,130]$;
    
    \item \emph{Coating Roll Gear Ratio Setpoint:} gap multiplier in the coating process, measured as a percentage. This parameter assumes discrete values in $[100,150]$;
    
    \item \emph{AM Composition:} composition of the slurry as a percentage of active material. This parameter is fixed at $96$.
\end{itemize}

Moreover a target variables is sampled from $800$ spatial locations, every eight seconds in the course of the three hours of the experiment:
\begin{itemize}
    \item \emph{Mass loading:} mass loading of the coating, measured in $\left[ \frac{g}{m^2}\right]$. This variable assumes continuous values in $[0,275)$;
\end{itemize}

From these recording, we build a dataset performing the following steps:
\begin{enumerate}
    \item First of all, we convert the recordings (\emph{Comma Bar Operator Position Actual}, \emph{Mass Loading}) to the same unit of measure used in the LRCS dataset.
    
    \item We compute a unique \emph{Comma Gap} measure as a product of \emph{Comma Bar Operator Position Actual} and \emph{Coating Roll Gear Ratio Setpoint}. This transformation is based on specific knowledge about the meaning of the parameters for the WMG coater machine. 
    
    \item We subselect the recordings in time; within the three hours of the experiment only short spans of time have a \emph{Comma Gap} actually set to an experimental value; most of the time the \emph{Comma Gap} variable is simply left to zero; we drop all the recordings when the comma gap is zero or when it is just transitioning to an experimental value. At the end, we retain those timesteps when the \emph{Comma Gap} variable is set to an experimental value, together with the corresponding values of \emph{Mass Loading}.
    
    \item We subselect the recordings in space; although mass loading is measured from 800 locations, this information is redundant; instead we consider the 100 central locations (which provide the most reliable measurements) and we average them into $n_{loc} = 2$ spatial measurements. 
\end{enumerate}

\subsubsection{Alignment of LRCS and WMG dataset}

The two datasets have a clear overlap and strong commonalities in their underlying models. In particular, they share a focus on modelling the casual relation between typical control parameters in the coating process and the resulting mass loading.

We then aim at setting up an abstraction between the underlying models. In order to do this, we will consider the WMG data and its associated model as the low-level model; this is justified by higher spatial resolution of the data (mass loading is measured at multiple locations) and the higher variable resolution for comma gap. Consequently, we will consider the LRCS data and its associate model as the high-level model.

However, before being able to define SCMs and setup a proper abstraction, we still need to perform further pre-processing to properly align the two datasets. We perform the following operations:

\begin{enumerate}
    \item From the WMG dataset we drop the control variable \emph{AM Composition}. Since it takes only a single value, it does not bring any information.
    
    \item From the LRCS dataset we drop the control variables \emph{S-to-L ratio} and \emph{Viscosity} which are related to the slurry preparation and not to the coating process.
    
    \item From the LRCS dataset we subselect for \emph{AM Composition}$ = 96$ in order to be consistent with the setting of the WMG dataset.
    
    \item We restrict \emph{Comma Gap} values in the LRCS dataset to $\{75,100,200\}$, excluding the values $\{50,300,400\}$ which are far out of the range considered in the WMG dataset.
    
    \item We extrapolate new values for \emph{Comma Gap} at $75$ and $200$ in the WMG dataset using a Gaussian process with a linear kernel trained on all the available WMG data. We then define the set of \emph{Comma Gap} values as $\{75,110,150,170,180,200\}$

    \item For both datasets we discretize the values of \emph{Mass loading} using $n_{bins}=5$ uniform bins.
\end{enumerate}

At the end of this process, the LRCS dataset contains $64$ samples, while the WMG dataset is constituted of $239$ samples.

\subsubsection{LRCS SCM}
The model underlying the LRCS dataset will constitute our abstracted model. We define this SCM making minimal assumptions. The set of endogenous variables in $\mathcal{M}^{LRCS}$ and their associated sets are defined as follows:
\begin{center}	
	\begin{tabular}{ccc}
	    & $\mathcal{X}^{LRCS} = \{ CG, ML \}$ & \tabularnewline
		& $\mathcal{M}^{LRCS}[CG] = \{ 75,100,200 \} $ & \tabularnewline
		& $\mathcal{M}^{LRCS}[ML] = \{ 0,1,...,n_{bins} \} $ & \tabularnewline
	\end{tabular}
\end{center}

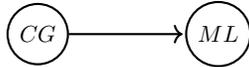
\begin{figure}[h]
    \begin{center}
    \begin{tikzpicture}[shorten >=1pt, auto, node distance=1cm, thick, scale=0.8, every node/.style={scale=0.8}]
    	\tikzstyle{node_style} = [circle,draw=black]
    	\node[node_style] (CG) at (0,0) {$CG$};
    	\node[node_style] (ML) at (3,0) {$ML$};
    	
    	\draw[->]  (CG) to (ML);
    \end{tikzpicture}
    \end{center}
    \caption{DAG of model $\mathcal{M}^{LRCS}$}
    \label{fig:LRCR_SCM}
\end{figure}

The DAG underlying the model $\mathcal{M}^{LRCS}$ is defined as in Figure \ref{fig:LRCR_SCM}

Notice that the edge simply expresses the physical causal dependence of \emph{Mass Loading} on the \emph{Comma Gap}. It is also worth noting that this SCM immediately represents the interventional setting in which the control parameter is regulated by an external experimenter.

Critically, as in most real-world scenario, we have no explicit knowledge of the mechanism that determines the variable $ML$ as a function of $CG$. This mechanism corresponds to a $n_{bins} \times 3$ matrix. We compute this matrix from observed frequencies: for each one of the three possible values of $CG$ we evaluate the empirical distribution of $ML$ into $n_{bins}$; at the end, we assemble these four empirical distributions into the mechanism matrix.

\subsubsection{WMG SCM}
The model underlying the WMG dataset will constitute our base model. We follow the same approach used above to define a SCM with minimal assumptions. The set of endogenous variables in $\mathcal{M}^{WMG}$ and their associated sets are defined as follows:
\begin{center}	
	\begin{tabular}{ccc}
	    & $\mathcal{X}^{WMG} = \{ CG, ML_1,..., ML_{n_{loc}} \}$ & \tabularnewline
		& $\mathcal{M}^{WMG}[CG] = \{ 75,110,150,170,180,200 \} $ & \tabularnewline
		& $\mathcal{M}^{WMG}[ML_i] = \{ 0,1,...,n_{bins} \} $ & \tabularnewline
	\end{tabular}
\end{center}

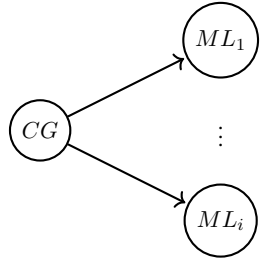
\begin{figure}[h]
    \begin{center}
\begin{tikzpicture}[shorten >=1pt, auto, node distance=1cm, thick, scale=0.8, every node/.style={scale=0.8}]
	\tikzstyle{node_style} = [circle,draw=black]
	\node[node_style] (CG) at (0,0) {$CG$};
	\node[node_style] (ML1) at (3,1.5) {$ML_1$};
	\node (dots) at (3,0) {$\vdots$};
	\node[node_style] (ML2) at (3,-1.5) {$ML_i$};
	
	\draw[->]  (CG) to (ML1);
	\draw[->]  (CG) to (ML2);
\end{tikzpicture}
\end{center}
    \caption{DAG of model $\mathcal{M}^{WMG}$}
    \label{fig:SCM_WMG}
\end{figure}

The DAG underlying the model $\mathcal{M}^{WMG}$ is defined as in Figure \ref{fig:SCM_WMG}.

Again, the edges simply express the physical causal dependence of \emph{Mass Loading} at different locations $i$ on \emph{Comma Gap}. 

This SCM implies $n_{loc}$ mechanisms, each one encoded into a $n_{bins} \times 6$ matrix. We apply the same approach used before to compute these matrix from the observed frequencies.

\subsubsection{Abstraction from the WMG model to the LRCS model}

Given the two SCMs above, we define a (partial) abstraction $\abs{}$ from $\mathcal{M}^{WMG}$ to $\mathcal{M}^{LRCS}$ as follows:
\begin{center}	
	\begin{tabular}{ccc}
		& $R=\left\{ CG, ML_i\right\}, \forall 1\leq i \leq n_{loc} $ & \tabularnewline
		& $a(CG)=CG$ & \tabularnewline
		& $a(ML_i)=ML, \forall 1\leq i \leq n_{loc}$ & \tabularnewline
	\end{tabular}
\end{center}

This correspond to the following abstraction:
\begin{center}
\begin{tikzpicture}[shorten >=1pt, auto, node distance=1cm, thick, scale=0.8, every node/.style={scale=0.8}]
	\tikzstyle{node_style} = [circle,draw=black]
	\node[node_style] (CG) at (0,0) {$CG$};
	\node[node_style] (ML) at (3,0) {$ML$};
	
	\node[node_style] (CGw) at (0,4) {$CG$};
	\node[node_style] (ML1) at (3,5.5) {$ML_1$};
	\node (dots) at (3,4) {$\vdots$};
	\node[node_style] (ML2) at (3,2.5) {$ML_i$};
	
	\draw[->]  (CGw) to (ML1);
	\draw[->]  (CGw) to (ML2);
	\draw[->]  (CG) to (ML);
	
	\draw[->,dashed]  (CGw) to (CG);
	\draw[->,dashed,bend left]  (ML1) to (ML);
	\draw[->,dashed,bend right]  (2.7,4) to (ML);
	\draw[->,dashed,bend right]  (ML2) to (ML);
\end{tikzpicture}
\end{center}

In the abstraction learning problem, we want to learn two weight matrices: the first one corresponds to $\alpha_{CG}$, and has shape $3 \times 6$; the second one corresponds to $\alpha_{ML}$ and has shape $n_{bins} \times (n_{bins})^{n_{loc}}$, in our case $5 \times 25$. Notice that, despite the limited number of averaging locations ($n_{loc}=2$) and discretizing bins ($n_{bin}=5$), the number of surjective functions to be evaluated in an enumeration algorithm is already unfeasible; indeed, using Equation \ref{eqn:all_n_surj_functions}, this value amounts to:
\begin{equation}
    |\mathcal{A}| = 4! \iiks{6}{4} \cdot 5! \iiks{25}{5} \approx 1.6 \cdot 10^{20}.
\end{equation}

\subsection{Experimental settings} \label{app:RealData_experimental}

We learn an abstraction between $\mathcal{M}^{WMG}$ and $\mathcal{M}^{LRCS}$ using our joint approach algorithm (see pseudocode in Appendix \ref{app:JointApproach}).

We run our algorithm considering all the possible combinations of hyperparameters in the following sets: temperature $T = \{0.1,0.2\}$, trade-off $\lambda = \{5.0,10.0\}$, and learning rate $\eta = \{0.001, 0.002, 0.005\}$. These values are chosen based on the previous experience on the synthetic simulations. For each setting we run an ensemble of $50$ models. Each model is trained for a number of epochs $n_e = 10^4$ and optimized by gradient descent using the Adam algorithm \cite{kingma2014adam}. At the end, we select the abstraction achieving the best JSD loss $\mathcal{L}_1$.

We evaluate the quality of the result in two ways. 
First, we discuss qualitatively the solution; although we do not have a ground truth, we can still comment on the pattern of the learned matrix comparing our expectations with the results of learning.
Second, we assess quantitatively whether transporting WMG data to the LRCS format, and integrating them with the existing LRCS data, may improve predictions. To do this, we set a three regression tasks with three different setups:
\begin{enumerate}
    \item \emph{(a) LRCS only:} we consider only the LRCS data. For each value $cg$ of the $CG$ variable, we train a regression model on all the LRCS samples for which $CG \neq cg$. We then test the model on the LRCS samples for which $CG=cg$.
    
    \item \emph{(b) LRCS plus WMG providing support:} we consider the LRCS and the transported $\abs{}(\textrm{WMG})$ data. For each value $cg$ of the $CG$ variable, we train a regression model on all the LRCS samples for which $CG \neq cg$ together with all the samples from $\abs{}(\textrm{WMG})$. We then test the model on the LRCS samples for which $CG=cg$.
    
    \item \emph{(c) LRCS plus WMG not providing support:} we consider the LRCS and the transported $\abs{}(\textrm{WMG})$ data. For each value $cg$ of the $CG$ variable, we train a regression model on all the LRCS and $\abs{}(\textrm{WMG})$ samples for which $CG \neq cg$. We then test the model on the LRCS and $\abs{}(\textrm{WMG})$ samples for which $CG=cg$.
\end{enumerate}

Task represents a realistic scenario in which a limited amount of data is used to infer a model that allows us to perform interpolation and extrapolation. Task (b) corresponds to a favorable case in which using abstracted data from another research group provides samples of control variables over which we are interpolating or extrapolating. Task (c) represents instead a harder scenario in which the abstracted data provides more information about the domain of interest, but has no samples for the specific values of the control variable for which we want to perform inference.

Notice that although our target variable is ordinal, we avoid using standard ordinal regression models as they can rarely deal with one value of the target variable being absent from the training data. Instead, since our target variable has been generated through uniform binning, we simply rely on a simple linear regression model with lasso penalty.

\subsection{Additional results}\label{app:RealData_additional}

\begin{figure}
\centering
\begin{minipage}[t]{0.29\linewidth}\centering
\resizebox{\columnwidth}{!}{%
\includegraphics{img/realexp_mat_cg.png}}
\caption{Learned $W_{CG}$}
\label{fig:app_realexp_mat_cg}
\end{minipage}\hfill%
\begin{minipage}[t]{0.69\linewidth}\centering
\resizebox{\columnwidth}{!}{%
\includegraphics{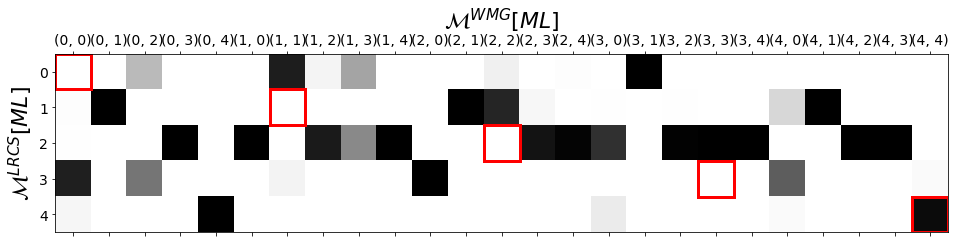}}
\caption{Learned $W_{ML}$}
\label{fig:app_realexp_mat_ml}
\end{minipage}
\end{figure}

Figure \ref{fig:app_realexp_mat_cg} shows the matrix $W_{CG}$ learned by the joint algorithm. The pattern of this matrix has been described in the main text. It is worth to point out that, formally, other binary matrices could be learned by our algorithm. A permutation of the values in the matrix $W_{CG}$, accompanied by another permutation in $W_{ML}$, could still achieve a very low loss value. Our expectation is mainly led by the semantics of the values to be mapped from the base model to the abstracted model: we expect identical values to be mapped to each other. Although this solution may indeed be the global optimum, the algorithm may stop in a local optimum with a vary low abstraction error which counter-intuitively maps values from the base model onto values in the abstracted model.

Figure \ref{fig:app_realexp_mat_ml} shows the matrix $W_{ML}$ learned by the joint algorithm. The pattern of this matrix is more difficult to interpret. This follows from a couple of considerations. First, while $CG$ is the variable we intervene upon, $ML$ is the variable we observe conditioned on the intervention. The mapping $W_{ML}$ has then to account for the mechanisms in $\mathcal{M}^{WMG}$ and $\mathcal{M}^{LRCS}$, making an intuitive mapping (highlighted again by the red border) less likely. In other words, the matrix $W_{ML}$ is first of all the matrix that minimizes the JSD and makes the abstraction diagram as commutative as possible; as soon as some noise is introduced in the mechanisms (as it is in our approximate case), the matrix $W_{ML}$ is affected.
Second, many values in the domain given by the Cartesian product of measurement at the two spatial locations in the base model are never realized (e.g.: a measurement corresponding to (0,4)); such values end up being unconstrained and they can assume any value as they do not affect the JSD loss of the learning algorithm.

\end{document}